%% file: main.tex
\documentclass{article}

\newif\ifarxiv
\arxivtrue

\ifarxiv
  \usepackage[preprint]{neurips_2026}
\else
  \usepackage[final]{neurips_2026}
\fi

\usepackage[utf8]{inputenc}
\usepackage[T1]{fontenc}
\usepackage{hyperref}
\usepackage{url}
\usepackage{booktabs}
\usepackage{amsfonts}
\usepackage{nicefrac}
\usepackage{amsmath}
\usepackage{microtype}
\usepackage{xcolor}

\newcommand{\guillaume}[1]{\textcolor{black}{#1}}

\usepackage{listings}
\lstdefinestyle{viperprompt}{
  basicstyle=\ttfamily\scriptsize\color{black},
  breaklines=true,
  columns=fullflexible,
  keepspaces=true,
  frame=single,
  framesep=4pt,
  aboveskip=6pt,
  belowskip=6pt,
  showstringspaces=false,
  upquote=true,
}
\usepackage{graphicx}
\usepackage{multirow}
\usepackage{subcaption}
\usepackage{makecell}
\usepackage{wrapfig}

\newcommand{\nquestions}{1,251}
\newcommand{\nimages}{419}
\newcommand{\norgans}{seven}          %
\newcommand{\nmcq}{419}
\newcommand{\nkprim}{414}
\newcommand{\nfreetext}{418}
\newcommand{\nmodels}{16}

\newcommand{\pathchatoverall}{51.0}
\newcommand{\toxscribepval}{<0.001}  %
\newcommand{\toxscribepvalgemma}{0.13}  %

\newcommand{\toxscribeqwenoverall}{62.4}

\newcommand{\toxscribegemmaoverall}{61.2}

\newcommand{\gemmavanillaoverall}{54.4}
\newcommand{\gptoverall}{56.0}

\newcommand{\toxqwen}{\textsc{ToxScribe}_{\text{Qwen}}}
\newcommand{\toxgemma}{\textsc{ToxScribe}_{\text{Gemma}}}
\newcommand{\tox}{\textsc{ToxScribe}}

\input{results/agreement_vars.tex}
\input{results/additional_vars.tex}

\title{VIPER: An Expert-Curated Benchmark for Vision-Language Models in Veterinary Pathology}

\author{
  Luca L. Weishaupt$^{1,2,3}$ \quad
  Simone de Brot$^{4}$ \quad
  Javier Asin$^{5}$ \quad
  Llorenç Grau-Roma$^{4}$ \\  
  \textbf{Nic G. Reitsam}$^{2,6}$ \quad
  \textbf{Andrew H. Song}$^{7}$ \quad 
  \textbf{Dongmin Bang}$^{2,3}$ \\ 
  \textbf{Stefan T. Kaluziak}$^{2,3}$ \quad
  \textbf{Long Phi Le}$^{2}$ \quad 
  \textbf{Jakob Nikolas Kather}$^{8}$ \\
  \textbf{Faisal Mahmood}$^{2,3,\dagger}$ \quad
  \textbf{Guillaume Jaume}$^{9,\dagger}$ \\
  $^{1}$ Harvard-MIT HST \quad
  $^{2}$ Mass General Brigham \quad
  $^{3}$ Harvard Medical School \\
  $^{4}$ COMPATH, University of Bern \quad
  $^{5}$ UC Davis \quad
  $^{6}$ University of Augsburg \\
  $^{7}$ UT MD Anderson Cancer Center \quad
  $^{8}$ TU Dresden \quad
  $^{9}$ University of Lausanne \\
  $^{\dagger}$ Co-senior authors \\ 
  \texttt{faisalmahmood@bwh.harvard.edu} $\quad$ \texttt{guillaume.jaume@unil.ch}
}

\begin{document}

\maketitle

\begin{abstract}
Pathology vision-language models are advancing rapidly, yet existing benchmarks remain focused on human tissue, particularly oncology, leaving non-human pathology largely unaddressed. This gap is especially important in toxicologic pathology, where microscopic tissue examination of laboratory animals is a core component of preclinical drug safety assessment. To address it, we introduce VIPER, the first expert-curated benchmark for vision-language model evaluation in toxicologic pathology. VIPER contains \nquestions{} questions associated with \nimages{} H\&E-stained rat histology images across \guillaume{seven organ systems}, covering multiple-choice, KPrim, and free-text formats. All questions were curated and validated by board-certified veterinary pathologists. In total, we benchmarked \nmodels{} models, including two newly introduced veterinary-pathology models, seven human pathology-specialized models, and seven general-purpose frontier models. The results identify a substantial domain gap between veterinary and human pathology, expose the risk of over-diagnosis of normal tissue in frontier models, and show that domain-specific training remains critical for visually grounded predictions. VIPER data and evaluation code are available at \url{https://github.com/mahmoodlab/viper}.
\end{abstract}

\section{Introduction}

Progress in AI for pathology has been driven by the availability of large public datasets and the development of standardized benchmarks. In human pathology, resources such as TCGA~\cite{weinstein2013tcga}, PANDA~\cite{bulten2020automated,bulten2022panda} or CAMELYON~\cite{litjens20181399} have supported a wide range of tasks, from diagnosis and grading to molecular prediction and segmentation~\cite{porpoise,wagner2023transformer,graham2019hover}. These efforts have in turn enabled broader benchmarks such as EVA~\cite{kaikoai2024eva} or HEST~\cite{jaume2024hest}, together with public leaderboards that make model comparison increasingly systematic. As a result, AI in pathology now benefits from a comparatively mature evaluation ecosystem relative to other medical imaging domains.

This progress, however, has been mostly centered on human clinical specimens, with a strong emphasis on oncology (e.g., TCGA consists entirely of tumor samples). Yet histopathology also plays a central role outside clinical care, particularly in veterinary and toxicologic pathology~\cite{seyhan2019lost}. In preclinical safety studies, pathologists examine animal tissue to characterize the drug toxicity and act as gatekeepers for whether a candidate compound can safely advance to human testing~\cite{cook2014lessons,waring2015analysis}. These workflows routinely involve large numbers of slides spanning multiple organs and dose groups, placing a substantial review burden on toxicologic pathologists.

Toxicologic pathology is therefore a natural target for AI, where the scale and repetitiveness create an opportunity for models that can assist with tissue screening and lesion reporting~\cite{turner2020society}. At the same time, progress in this domain has remained constrained by the lack of public benchmarks. Existing veterinary pathology datasets are few and often focused on isolated tasks such as mitosis detection or tumor classification~\cite{wilm2022catch,kumar2020deep}. The gap is even more pronounced for vision-language reasoning. While pathology vision-language models have advanced rapidly in recent years, nearly all have been developed and evaluated in human pathology, and often on educational or web-derived resources that emphasize clinical oncology and textbook-style question answering~\cite{pathchat,quilt_llava,sunPathGen16M16Million2024,sun2024pathmmu}. In non-human pathology, vision-language evaluation remains essentially absent.

This gap raises three central questions. How well do human pathology models transfer to veterinary pathology? How do general-purpose frontier models behave on pathology vision-question answering? And what is gained by training models directly on veterinary pathology data? Here, we introduce VIPER, an expert-curated benchmark for evaluating vision-language models in rodent toxicologic pathology. VIPER contains \nquestions{} questions associated with \nimages{} H\&E-stained rat histology images spanning \guillaume{\norgans{} organ systems}. It includes multiple-choice, KPrim, and free-text formats. All questions were authored and validated by veterinary pathologists board-certified by the European College of Veterinary Pathologists (ECVP). All questions were anchored in tissue morphology, then refined through human-in-the-loop generation, and adversarially filtered to reduce hallucination and ``mirage predictions''~\cite{asadi2026mirage} (i.e., answers driven by textual priors instead of visual understanding). We further introduce $\tox$, two veterinary pathology-specialized vision-language models built on Qwen3.5-27B and Gemma 4 backbones.

We make four main contributions. First, we introduce VIPER, to our knowledge, the first benchmark for vision-language model evaluation in non-human pathology, focused on rat toxicologic histology.
Second, we introduce a human-in-the-loop benchmark construction pipeline centered on morphology-grounded question design, expert refinement, and adversarial filtering to generate high-quality question-answer pairs.
Third, we provide an in-depth reader study and ablation experiments to show that VIPER supports consistent expert judgments.
Fourth, we benchmark \nmodels{} models, including veterinary pathology-specialized models, human pathology-specialized models, and frontier general-purpose models.

\begin{figure}[t]
    \centering
    \vspace{0.5em}
      \includegraphics[width=\textwidth]{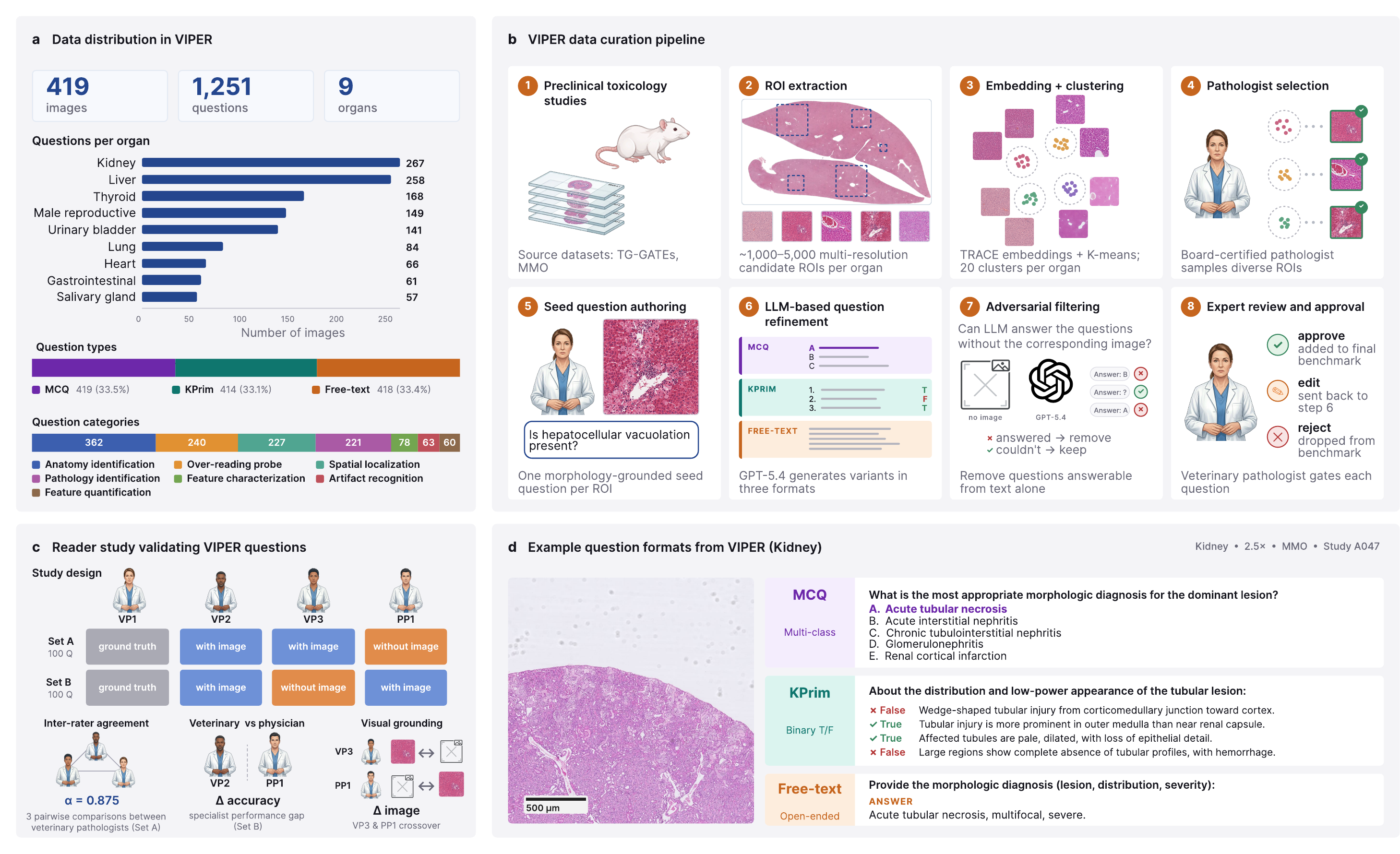}
\caption{
    \textbf{VIPER benchmark overview.} 
    \textbf{(a)} VIPER contains 419 H\&E-stained rat histology images from \guillaume{\norgans{} organ systems} and 1,251 questions spanning multiple-choice (MCQ), KPrim, and free-text. 
    \textbf{(b)} Overview of the proposed expert-guided data curation pipeline.
    \textbf{(c)} Reader study set up with three veterinary pathologists certified by the European College of Veterinary Pathologists (ECVP), denoted VP1--VP3, and a board-certified physician pathologist (PP1) trained in human pathology.
    \textbf{(d)} Example of questions from VIPER.
    VP: Veterinary Pathologist, PP: Physician Pathologist. 
    }
    \label{fig:figure1}
\end{figure}

\section{Related Work}

\paragraph{Vision-language models in pathology.} Vision-language modeling in pathology has developed mainly along two lines. First, contrastive image-text models that are trained to align pathology images and captions in a shared embedding space, often using domain-specific vision encoders~\cite{plip,conch,uni,virchow,xuWholeslideFoundationModel2024}. Second, generative and instruction-tuned models that have adapted these backbones for open-ended question answering and conversational assistance~\cite{pathchat,zhang2025patho}. Across both lines, recent work has increasingly scaled training with large collections of scientific figures, textbooks, educational websites, and teaching videos, often relying on automatically constructed image-text or question-answer pairs~\cite{quilt,quilt_llava,sunPathGen16M16Million2024}. Together, these efforts have produced pathology-specialized models that, in some settings, can surpass general-purpose multimodal frontier models~\cite{pathchat}. Nearly all prior work, however, has focused on human pathology, often with a strong emphasis on oncology, and has been evaluated on benchmarks derived from human educational and scientific material that typically cover only a narrow range of organs and cancer types. This leaves two open questions: whether these models transfer to non-human pathology, and whether these models are truly capable of image-grounded reasoning.

\paragraph{Benchmarking of vision-language models.} Pathology vision-language models are most often evaluated on open-ended question-answer benchmarks. PathMMU~\cite{sun2024pathmmu} is the reference benchmark dataset in this space, and is composed of multiple-choice questions with explanations. Earlier resources such as PathVQA~\cite{he2020pathvqa} pair pathology images with question-answer data derived from textbooks and the PEIR library, while QuiltVQA~\cite{quilt_llava} is constructed from pathology teaching videos. More broadly, publicly available pathology VLM datasets have been generated  (semi-)automatically from existing image-caption pairs (e.g., Quilt-1M~\cite{quilt}) with limited oversight from pathologists in designing the questions-answer pairs. In addition, in some cases, the questions-answers were synthesized by now-obsolete general-purpose models (e.g., GPT4-V) prone to hallucination. Finally, because many pathology QA datasets are derived from educational material, models may exploit stylistic cues or recurring patterns instead of truly reasoning about the tissue morphology.

\paragraph{Veterinary pathology and toxicologic pathology.} Veterinary pathology involves the microscopic evaluation of animal tissues outside the human clinical setting. Rodent histology is of particular importance due to its use in research labs and in toxicologic pathology, which plays a central role in preclinical safety assessment \cite{cook2014lessons,seyhan2019lost,waring2015analysis}. In drug safety studies, pathologists examine tissue sections from animal studies to identify treatment-related lesions and document abnormalities across all major organ systems, often under increasing dose levels. \textit{In vivo} non-human toxicologic pathology is central for drug development, as no therapeutic agent can enter human testing without prior preclinical safety evaluation~\cite{fda2024glp}. Compared with clinical pathology benchmarks, this setting is distinguished by systematic multi-organ sampling, very high volumes of slides per study (up to 20,000 tissue slides for 2-year carcinogenicity studies), and a predominance of non-neoplastic findings. The dominant morphologic patterns are degeneration, necrosis, inflammation, hypertrophy, vacuolation, and adaptive change. In practice, pathologists must interpret these findings in the context of drug administration dose and duration, as well as species and background strain. 

\paragraph{Deep learning for toxicologic pathology.} Deep learning in toxicologic pathology remains at an early stage, with prior work focusing mainly on narrowly defined tasks such as lesion detection and quantification, normal tissue recognition, and organ identification~\cite{kuklyte2021evaluation,holger2021histonet,citlalli2022mmonet}. These studies suggest that deep learning models can support key components of toxicologic histopathology workflows, but also highlight a field that is still limited in model diversity and evaluation compared to human pathology~\cite{mehrvar2021deep}. Recent perspective pieces point to growing interest in applying AI across toxicology, while emphasizing methodological and translational challenges~\cite{klambauer2023introduction,mehrvar2021deep}. Large-scale representation learning using self-supervised learning has only just begun to emerge, with only one foundation model reported so far for non-human pathology~\cite{jaume2024deep}. Vision-language modeling, however, remains essentially unexplored in this domain, as do curated vision-language benchmarks. \guillaume{In broader veterinary pathology, public datasets are centered on companion animals~\cite{aubreville2023midog,wilm2022catch}.} 

\section{VIPER: Veterinary Pathology Image Evaluation and Reasoning}
\label{sec:viper}

\subsection{Overview}

VIPER is an expert-curated benchmark for vision-language model evaluation in rodent toxicologic pathology. It comprises \nquestions{} question-answer pairs associated with \nimages{} H\&E-stained regions of interest from rat (\textit{Rattus norvegicus}) preclinical toxicology studies (\textbf{\autoref{fig:figure1}.a}). \guillaume{VIPER covers seven organ systems, including the male reproductive system ($n=50$ images), the digestive system ($n=40$), the urinary system ($n=137$), the hepatobiliary tract ($n=86$), the endocrine system ($n=56$), the respiratory system ($n=28$), and the cardiovascular system ($n=22$). At the structure level, VIPER covers 23 distinct named anatomic structures (\textbf{Appendix~\autoref{tab:structures}}). 
} Organ systems were selected based on image availability and on whether they contained pathologic lesions or relevant anatomic structures to support morphology-grounded questions. All annotations were validated by board-certified veterinary pathologists with expertise in rodent histology. To test different forms of multimodal reasoning, VIPER spans three question formats: multiple-choice questions (MCQ), KPrim questions composed of multiple true/false statements, and free-text open-ended questions (\textbf{\autoref{fig:figure1}.d}). In total, the benchmark contains \nmcq{} MCQs, \nkprim{} KPrim questions, and \nfreetext{} free-text questions \guillaume{with 63 images sourced from TG-GATEs\cite{igarashi2014open} (189 questions) and 356 from MMO\cite{citlalli2022mmonet} (1,062 questions).} Each question is linked to a 1,024$\times$1,024-pixel ROI at one of several magnifications with 304 images at 20$\times$ (0.5~$\mu$m/px), 54 images at 5$\times$ (2~$\mu$m/px), and 61 images at 2.5$\times$ (4~$\mu$m/px). Reflecting the composition of preclinical safety tissue, a substantial fraction of VIPER assesses normal anatomy and whether tissue is within normal limits.

\subsection{Expert-driven annotation pipeline}
\label{sec:annotation}

We developed a web app to support image visualization, ROI selection, question authoring, and peer validation. The interface was designed to keep the full curation workflow in one place (\textbf{\autoref{fig:figure1}.b}).

\paragraph{Image selection.} First, a large pool of candidate ROIs was randomly extracted for each organ ($\approx$1{,}000 to 5{,}000 ROIs per organ). \guillaume{ROIs were randomly positioned crops of the whole-slide image, so a focal finding can appear wherever in the image.}  These candidates were drawn from small-molecule preclinical toxicology studies sourced from TG-GATEs~\cite{igarashi2014open} (157 studies, CC BY-SA 2.1 JP license) and MMO~\cite{citlalli2022mmonet} (9 studies, CC BY-NC 4.0 license). To facilitate review and ensure morphologic diversity of the selected ROIs, candidates were embedded using the TRACE vision encoder~\cite{jaume2024deep} (trained on TG-GATEs) and K-means clustered into 20 bins within each organ. The pathologist then sampled images from each bin, ensuring coverage of a broad range of tissue types and diverse histologic morphologies. 

\paragraph{Seed question authoring.} For each selected ROI, a board-certified veterinary pathologist wrote one seed question answerable from the tissue morphology. For example, \textit{``Where is the most pronounced (artefactual) atelectasis?''} with answer \textit{``bottom center.''} These seed questions were designed to probe multiple forms of diagnostic reasoning, including anatomy and pathology identification, normal assessment, and artifact detection. The goal was to anchor each sample in a visually meaningful question that could not be answered reliably without inspecting the image.

\paragraph{Human-in-the-loop question refinement.} Starting from the seed questions, we used an LLM \guillaume{(GPT-5.2)} to generate variants spanning the three question formats: (i) multiple-choice questions (five answer options each) to test multi-class classification, (ii) KPrim questions (four true/false statements each) to test binary classification, and (iii) free-text questions to test open-ended reasoning. To reduce hallucination and mirage predictions~\cite{asadi2026mirage} (i.e., correct answers driven by text priors or dominant-answer guessing rather than visual evidence), we designed an adversarial filter. Specifically, for each MCQ and KPrim candidate, we queried GPT-5.2 (temperature 0) with the question stem and answer options but no image, repeating each query three times with the MCQ option order reshuffled per trial. A candidate was flagged guessable if any MCQ trial was correct (strict policy) or if the worst-case KPrim trial answered $\geq 3/4$ statements correctly. All flagged candidates were regenerated by \guillaume{GPT-5.2} with explicit feedback as part of the prompt up to three times before being escalated to a pathologist for manual revision or removal of the seed question.
\guillaume{The regeneration prompt includes the failed stem, its answer and distractors, and the guesser's own stated reasoning for why it could guess. All prompts are reproduced in \textbf{Appendix~\ref{app:prompts}}.} This process led to 247 of the 708 seed-question images (34.9\%) being dropped. \guillaume{In total, the authoring pathologist contributed around 80 hours, corresponding to roughly 4 minutes of board-certified veterinary pathologist time per released question and about 11 minutes per released image.} Free-text variants were not adversarially filtered, as they reformulate the seed Q\&A directly and were validated by the authoring pathologist. Each free-text question was also paired with an LLM-generated scoring rubric, reviewed alongside the question by the authoring pathologist and used at evaluation time. All final questions were reviewed by a veterinary pathologist, who manually approved, revised, or rejected each question.

\paragraph{Question categorization.} Every question was further split into one of seven categories, including anatomy identification ($n=362$), an ``over-reading'' probe (check for lesion hallucinations on normal tissue, $n=240$), spatial localization ($n=227$), pathology identification ($n=221$), feature characterization ($n=78$), artifact recognition ($n=63$), and feature quantification ($n=60$). Categories were assigned at the seed question-level and inherited by the three reformulations (MCQ, KPrim, free text). Full definitions as well as examples are provided in \textbf{Appendix~\autoref{tab:taxonomy-definitions}}.

\subsection{Evaluation protocol}
\label{sec:eval-protocol}
 
The benchmark performance was evaluated separately for each question format and then aggregated into an overall score. Multiple-choice questions were scored using exact match, assigning a score of 1 to the correct option and 0 otherwise (similar to an accuracy score). For each question, we include five candidate answers, giving an expected random baseline of 0.20. To measure each model's robustness to answer-position bias, every MCQ is presented in five cyclic-shift permutations of the answer ordering, and the reported MCQ accuracy is the mean across permutations. KPrim questions were scored using a 4-statement true/false question format with the ETH ``half-point'' rule (4/4 correct $\rightarrow$ 1.0, 3/4 $\rightarrow$ 0.5, $\leq$2/4 $\rightarrow$ 0.0), giving an expected random baseline of 0.1875~\cite{krebs1997swiss}. \guillaume{Two thirds of the benchmark is therefore scored deterministically.} Free-text responses were evaluated using GPT-5.4 as an LLM judge under a calibrated two-axis rubric that weighted diagnostic accuracy at 70\% and completeness at 30\%, with the per-question scoring rubric included in the judge prompt as additional context. To ensure the consistency of the LLM judge, we conducted preliminary calibration experiments, where we found that the free-text ranking of $\toxqwen$ over PathChat+ was preserved under both a repeated GPT-5.4 and Claude Opus 4.7 as judges, with a gap of $5.4$-$5.7$\% ($p \le 0.010$, paired bootstrap, \textbf{Appendix~\ref{app:judge-robustness}}). \guillaume{Item-level scores from the two judges are nearly collinear (Pearson $r=0.98$, Spearman $\rho=0.95$, $n=1{,}254$ paired predictions) and a repeated call to the same judge gives $r=0.99$.}

\subsection{Expert reader agreement on VIPER}
\label{sec:reader-study}

To validate the quality of the resulting samples, we ran a reader study involving three ECVP board-certified veterinary pathologists: $\mathrm{VP}_1$, the benchmark author and gold standard, and two external readers, $\mathrm{VP}_2$ and $\mathrm{VP}_3$ (\textbf{\autoref{fig:figure1}.c}). We randomly sampled 100 image-question pairs from VIPER, denoted Set~A, and asked the external readers to answer each item through a custom online platform. Agreement was quantified with Krippendorff's $\alpha$, where $\alpha=0$ corresponds to chance agreement and $\alpha=1$ to perfect agreement. The results show strong concordance across veterinary experts, with Krippendorff's $\alpha=\krippAMCQ{}$ on MCQ and $\alpha=\krippAKPrim{}$ on KPrim questions. This high agreement supports the validity of VIPER as an expert-curated toxicologic pathology benchmark that can be answered consistently by domain specialists. A second 100-question Set~B, introduced below, expands the study to a physician pathologist and a no-image condition. Additional reader study results are reported in \textbf{Appendix~\autoref{tab:interrater}, \autoref{tab:multi-rater-reliability} and \autoref{tab:ft-icc}}.

\section{Baselines}
\label{sec:baselines}

We evaluated \nmodels{} models spanning three categories: (i) veterinary pathology-specialized VLMs, (ii) human pathology-specialized VLMs, and (iii) generic-purpose multimodal frontier models. All models are evaluated under the same prompting and generation protocol. 

\paragraph{Veterinary-specialized models.} Because no VLMs currently exist for non-human pathology, we developed two veterinary-specialized models for this study. The first variant, denoted $\toxqwen$, is based on a Qwen3.5-27B multimodal LLM~\cite{qwen35_modelcard}, and the second variant, denoted $\toxgemma$, is based on a Gemma~4 multimodal LLM~\cite{gemma4_modelcard}. Both models were trained on the same instruction-tuning set, composed of approximately 345{,}000 pathology ROIs and approximately 4 million instruction pairs derived from image captions. Training data were assembled through a multi-step pipeline built from permissively reusable sources (e.g., PDFs from PubMed Open Access, regulatory agency documents, and teaching materials). The resulting corpus spans all major organ systems across multiple species, with a large proportion of rodent tissue. Both models were instruction-tuned with LoRA adapters while keeping the vision backbone frozen. Additional training and preprocessing details are provided in the \textbf{Appendix~\ref{app:toxscribe-training}}. Importantly, $\tox$ was not trained on whole-slide images or ROIs derived from TG-GATEs and MMO, and all questions in VIPER are novel and do not appear in any public dataset, ensuring no training data leakage for $\tox$ (or any other evaluated model).

\paragraph{Human pathology-specialized models.} We evaluated pathology-specialized VLMs developed primarily for human histopathology, including PathChat+~\cite{pathchat,weishaupt2025evidence}, Patho-R1-7B~\cite{pathor125}, Patho-R1-3B~\cite{pathor125}, MedGemma-1.5-4B~\cite{sellergren2025medgemma}, QuiltLLaVA~\cite{quilt_llava}, PathGenLlaVa~\cite{sunPathGen16M16Million2024}, and LlaVaMed~\cite{llava-med}. These models differ in architecture and training data, but all have been developed with substantial exposure to human medical imaging. In particular, PathChat+ was trained on in-house annotations and PubMed Central, MedGemma-1.5-4B was trained on pathology together with radiology, dermatology, and ophthalmology images, QuiltLLaVA was tuned for histopathology using localized narratives extracted from open-source pathology videos, PathGenLlaVa was built on large-scale synthetic pathology image-text pairs generated from whole-slide images, and LlaVaMed was developed as a broader biomedical vision-language assistant. Patho-R1-7B, Patho-R1-3B, MedGemma-1.5-4B, QuiltLlaVa, PathGenLlaVa, and LlaVaMed were evaluated from their official HuggingFace or public releases. PathChat+ was obtained directly from the authors. 

\paragraph{General-purpose multimodal frontier models.} We evaluated both closed-weight and open-weight generic-purpose multimodal models. Closed-weight commercial models accessed via API include GPT-5.4 (March 5, 2026), GPT-5.4-mini (March 17, 2026), GPT-5.4-nano (March 17, 2026), Claude Sonnet~4.6 (May 22, 2025), and Gemini 2.5 Flash (February 19, 2026). Open-weight models include Gemma~4-31B-it (April 2, 2026) and Qwen3.5-27B (February 24, 2026), with both also serving as the initialization for $\tox$, enabling a direct comparison of domain-specific fine-tuning to zero-shot performance. We report results for a single inference pass per question with a fixed temperature (\texttt{temperature=0}), without tool use, external retrieval, or web access.

\section{Results}
\label{sec:results}

\subsection{VIPER performance}

\input{results/main_results.tex}

\begin{figure}[t]
  \centering
  \includegraphics[width=\textwidth]{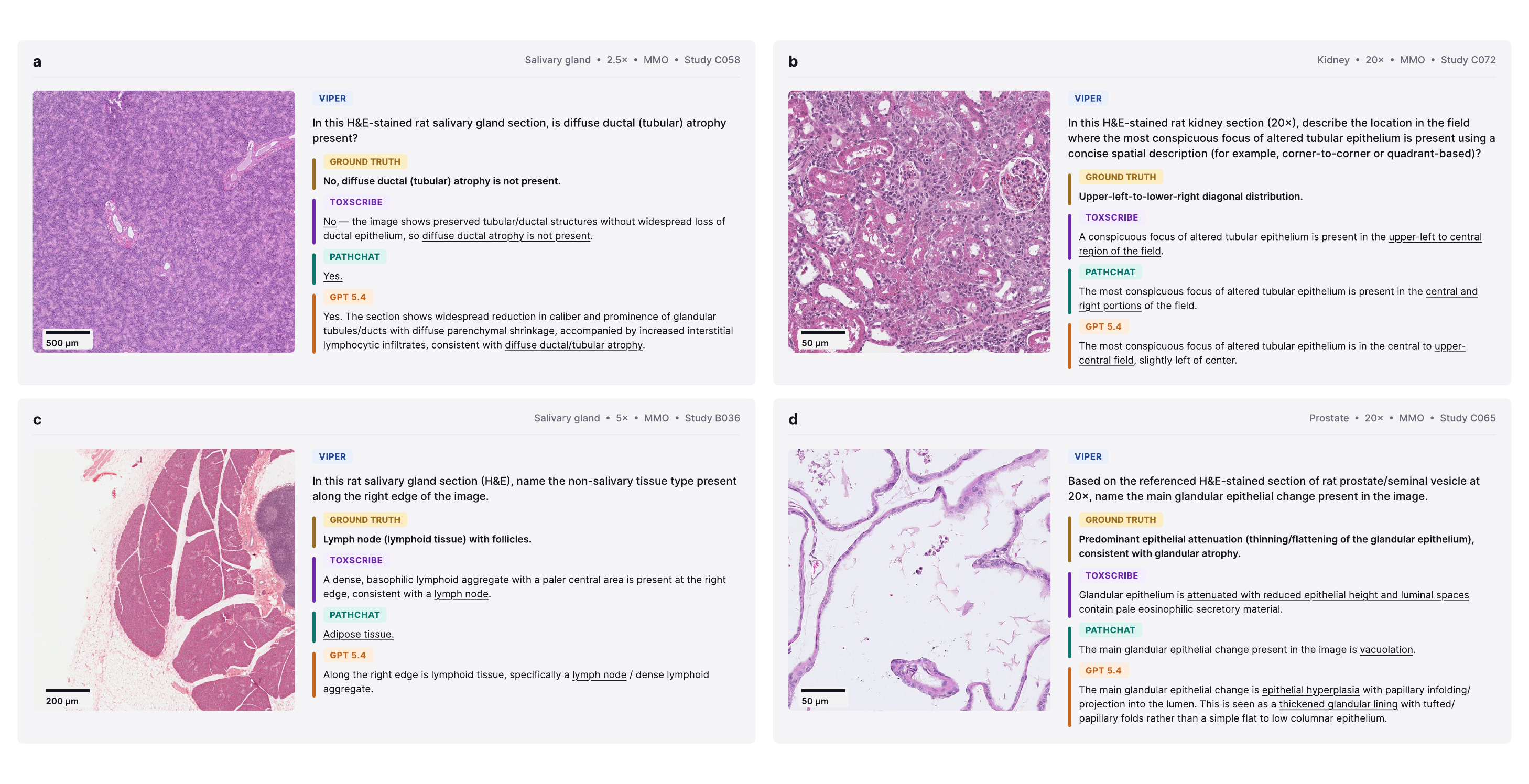}
  \caption{\textbf{Qualitative comparison of three model responses across representative questions in VIPER.} Each panel shows an H\&E image from rat tissue together with the benchmark question, the ground-truth answer authored by $\mathrm{VP}_1$ (used as the evaluation reference, see \autoref{sec:eval-protocol}), and responses from $\tox$, PathChat+, and GPT-5.4. The examples illustrate four of the seven VIPER question categories (\textbf{Appendix~\autoref{tab:taxonomy-definitions}}):
  \textbf{a} over-reading probe, distinguishing normal histomorphology from pathologic change in salivary gland,
  \textbf{b} spatial localization, identifying the in-image location of a focus of altered tubular epithelium in kidney,
  \textbf{c} anatomy identification, recognizing a non-salivary tissue type within a salivary gland section, and
  \textbf{d} pathology identification, naming the predominant glandular epithelial change in prostate / seminal vesicle. Underlined text marks the core answer content.
  }
  \label{fig:qualitative}
\end{figure}

\paragraph{Overall performance.} The main results of VIPER are presented in \textbf{\autoref{tab:main-results}} with qualitative examples reported in \textbf{\autoref{fig:qualitative}}. The two $\tox$ variants lead the benchmark, with $\toxqwen$ at \toxscribeqwenoverall{}\% and $\toxgemma$ at \toxscribegemmaoverall{}\% (both statistically similar ($P=\toxscribepvalgemma{}$, paired item-level bootstrap, 10{,}000 resamples), suggesting that domain-adaptation benefit transfers across backbones. $\toxqwen$ significantly outperforms every non-veterinary-specialized model ($P\toxscribepval{}$ for each model, \textbf{Appendix \autoref{tab:significance_categories}}). Human pathology-specialized models show a wide spread, ranging from 16.2\% (LLaVA-Med) to \pathchatoverall{}\% (PathChat+). Flagship frontier models from OpenAI, Anthropic, and Google range from 41.0\% to \gptoverall{}\%, with the strongest open-weight model (Gemma~4, \gemmavanillaoverall{}\%) close to the strongest closed-weight model (GPT-5.4, \gptoverall{}\%).

\paragraph{Performance by question category.} Performance differs markedly across the seven question categories (\textbf{Appendix~\autoref{tab:category_performance}}). The clearest separation occurs in the ``over-reading probe'' category, which tests whether models can resist hallucinating lesions when the image is within normal limits (\textbf{\autoref{fig:qualitative}.a}). $\toxqwen$ and $\toxgemma$ achieve 77.2\% and 80.5\%, respectively. The best non-veterinary model reaches only 62.1\% (PathChat+), and the leading frontier model only 55.1\% (GPT-5.4). This gap suggests that frontier models may be more susceptible to lesion overcalling, either because their training distributions over-represent pathologies, or because post-training objectives favor responses aligned with the prompt suggesting a lesion~\cite{ibrahim2026warm}. By contrast, $\tox$ was trained on both normal and abnormal morphologies using supervised instruction fine-tuning, without human-preference optimization, which may improve calibration. The ``Pathology identification'' category remains challenging for all models ($\toxqwen$: 50.3\%, GPT-5.4: 49.2\%, $\toxgemma$: 48.3\%), highlighting the complexity of distinguishing lesions. In the remaining categories, the two $\tox$ variants and GPT-5.4 perform within a few percentage points of each other.

\paragraph{Performance by organ \guillaume{system}.} \guillaume{Organ-level results show that $\tox$ advantage is consistent across most tissues (\textbf{Appendix~\autoref{tab:per-organ}}). The two $\tox$ variants rank first in six of the seven organ systems, with the largest gains in urinary (9.5\%), respiratory (6.0\%), endocrine (5.6\%), and hepatobiliary (5.1\%), and the narrowest margin in digestive (2.0\%). Compared to PathChat+ (best human pathology model), $\tox$ outperforms it in every organ system.}

\subsection{Insights into VIPER}
\label{sec:insights}

\begin{table}[h]
  \centering
  \small
  \caption{\textbf{Pathologist reader performance on VIPER by question format.} Scores are computed against the $\mathrm{VP}_1$ gold standard and reported as percent ($0$--$100$). MCQ is scored as percent correct. KPrim uses the ETH half-point rule (4/4 correct statements $\rightarrow 100$, 3/4 $\rightarrow 50$, $\leq 2/4 \rightarrow 0$) and the reported value is the mean across questions. Free-text is scored on $[0,100]$ by an LLM judge (\texttt{\ftJudgeModel{}} as $0.7\cdot$diagnostic accuracy + $0.3\cdot$completeness, both rated 0--10, mapped to percent), with the per-question scoring rubric included in the prompt. Bracketed values are 95\% bootstrap CIs ($10^4$ item-level resamples).}
  \label{tab:per-rater-acc}
  \begin{tabular}{llrlrlrl}
    \toprule
    & & \multicolumn{2}{c}{MCQ} & \multicolumn{2}{c}{KPrim} & \multicolumn{2}{c}{Free-text} \\
    \cmidrule(lr){3-4}\cmidrule(lr){5-6}\cmidrule(lr){7-8}
    Rater & Set & $n$ & Acc.\ [95\% CI] & $n$ & ETH [95\% CI] & $n$ & Score [95\% CI] \\
    \midrule
    $\mathrm{VP}_2$ & A & \accVPtwoSetAMCQn   & \accVPtwoSetAMCQ~[\accVPtwoSetAMCQcilo, \accVPtwoSetAMCQcihi]   & \accVPtwoSetAKPrimn   & \accVPtwoSetAKPrim~[\accVPtwoSetAKPrimcilo, \accVPtwoSetAKPrimcihi]     & \accVPtwoSetAFTn   & \accVPtwoSetAFT~[\accVPtwoSetAFTcilo, \accVPtwoSetAFTcihi] \\
    $\mathrm{VP}_2$ & B & \accVPtwoSetBMCQn   & \accVPtwoSetBMCQ~[\accVPtwoSetBMCQcilo, \accVPtwoSetBMCQcihi]   & \accVPtwoSetBKPrimn   & \accVPtwoSetBKPrim~[\accVPtwoSetBKPrimcilo, \accVPtwoSetBKPrimcihi]     & \accVPtwoSetBFTn   & \accVPtwoSetBFT~[\accVPtwoSetBFTcilo, \accVPtwoSetBFTcihi] \\
    $\mathrm{VP}_3$ & A & \accVPthreeSetAMCQn & \accVPthreeSetAMCQ~[\accVPthreeSetAMCQcilo, \accVPthreeSetAMCQcihi] & \accVPthreeSetAKPrimn & \accVPthreeSetAKPrim~[\accVPthreeSetAKPrimcilo, \accVPthreeSetAKPrimcihi] & \accVPthreeSetAFTn & \accVPthreeSetAFT~[\accVPthreeSetAFTcilo, \accVPthreeSetAFTcihi] \\
    $\mathrm{PP}_1$ & B & \accHPoneSetBMCQn   & \accHPoneSetBMCQ~[\accHPoneSetBMCQcilo, \accHPoneSetBMCQcihi]   & \accHPoneSetBKPrimn   & \accHPoneSetBKPrim~[\accHPoneSetBKPrimcilo, \accHPoneSetBKPrimcihi]     & \accHPoneSetBFTn   & \accHPoneSetBFT~[\accHPoneSetBFTcilo, \accHPoneSetBFTcihi] \\
    \bottomrule
  \end{tabular}
\end{table}

\paragraph{Veterinary vs. human pathology.} VIPER exposes a clear gap between human and veterinary pathology. The best human pathology-specialized baseline (PathChat+, \pathchatoverall{}\%), is significantly outperformed by $\toxqwen$ by 11.5\% ($P<0.001$). This establishes that training on human pathology data alone is insufficient for rat toxicologic pathology. To test whether this observation translates to human readers, we expanded our reader study by asking $\mathrm{VP}_2$ and a board-certified physician pathologist trained in human pathology ($\mathrm{PP}_1$) to independently answer a second 100-question set, Set~B. On MCQ, $\mathrm{VP}_2$ scored \meanVPvHPSetBMCQVP{}\% vs.\ $\mathrm{PP}_1$'s \meanVPvHPSetBMCQHP{}\% ($p=\pvalVPvHPSetBMCQ{}$, paired item-level bootstrap). KPrim ($\Delta=\deltaVPvHPSetBKPrim{}$\%, $p=\pvalVPvHPSetBKPrim{}$) and free-text ($\Delta=\deltaVPvHPSetBFT{}$\%, $p=\pvalVPvHPSetBFT{}$) gaps are not significant (\textbf{\autoref{tab:per-rater-acc}}, \textbf{Appendix \autoref{tab:vet-vs-human}}).
Lower free-text scores on Set~B mainly reflect borderline cases. In five of the ten low-scoring Set~B free-text items ($<30$\%), the reference answer was ``no significant histopathologic change'', whereas both external readers described subtle morphologic changes.

\paragraph{The importance of visual grounding.} To validate that VIPER truly requires the image rather than text priors alone, we re-evaluated $\toxqwen$, PathChat+, and GPT-5.4 by prompting the models with the questions without providing the corresponding image. Across the 1,251 VIPER questions, not providing images produces large drops on every model: $\toxqwen$ loses 41.7\% across all questions, PathChat+ loses 28.0\% and GPT-5.4 loses 29.0\% (all with $P<0.001$, paired item-level bootstrap). Additional results are shown in \textbf{Appendix \autoref{tab:ablation-matrix}}. We also expanded our reader study to compare $\mathrm{VP}_3$ and $\mathrm{PP}_1$ to themselves across their with- and without-image sets (\textbf{\autoref{tab:image-effect}}). Removing the image produces large significant drops on MCQ for both readers ($\mathrm{VP}_3$: $\Delta = \deltaImgVPthreeMCQ{}$\%, $p = \pvalImgVPthreeMCQ{}$, $\mathrm{PP}_1$: $\Delta = \deltaImgHPoneMCQ{}$\%, $p = \pvalImgHPoneMCQ{}$, unpaired bootstrap test). On KPrim, $\mathrm{VP}_3$ shows a comparable drop ($\deltaImgVPthreeKPrim{}$\%, $p < 0.001$), with a smaller non-significant drop for $\mathrm{PP}_1$ ($\deltaImgHPoneKPrim{}$\%, $p = \pvalImgHPoneKPrim{}$).

\paragraph{Gain from domain-specific fine-tuning.} Domain-specific fine-tuning yields a clear and statistically robust gain for both open-weight backbones. $\toxqwen$ improves by 10.0\% over the base Qwen3.5-27B model and $\toxgemma$ improves by 6.8\% over the base Gemma~4 model (paired item-level bootstrap, $10^4$ resamples, $P<0.001$ in both cases). These gains are substantial given the rapid progress of generic-purpose multimodal models, and show that targeted adaptation remains highly valuable when the task requires specialized visual knowledge. We emphasize that the training data for the veterinary pathology-specialized models are fully independent from VIPER, with no risk of data leakage. 

\paragraph{Challenges in understanding instructions.} Several models face difficulty in understanding instructions. The Patho-R1 family (Patho-R1-3B and Patho-R1-7B) is biased toward answering \texttt{True} on most KPrim statements (90.2\% and 87.6\% of statements, respectively), even though both perform competitively on free-text (36.1\% and 46.0\%, respectively). LLaVA-Med shows a parallel ``Option-A bias'' on MCQ, picking option~A on 67.5\% of questions despite the correct answer rotating uniformly across positions. These patterns suggest alignment issues introduced during fine-tuning, consistent with shortcut learning~\cite{geirhos2020shortcut} or insufficient instruction diversity.

\paragraph{Oracle performance.} Beyond aggregate scores, models also differ in which questions they answer correctly. The leading models, $\toxqwen$, PathChat+, and GPT-5.4 fail on largely different questions, indicating substantial complementarity across models. On MCQ at the permutation level ($n=2{,}095 = 419$ base questions $\times$ 5 cyclic-shift rotations), $\toxqwen$ is correct while GPT-5.4 is wrong on 20.4\% of items, whereas GPT-5.4 is correct while $\toxqwen$ is wrong on 11.7\% of items. On free-text ($n=418$), $\toxqwen$ uniquely ``rescues'' 4.8\% of questions (LLM judge score $>0$ for $\toxqwen$ while both GPT-5.4 and PathChat+ score 0), compared with 3.6\% rescued uniquely by GPT-5.4. Therefore, an ``oracle'' predictor that selects, for each question, the highest per-sample score among the three baselines achieves an overall score of 78.2\% (MCQ: 82.3\%, KPrim: 73.7\%, and free-text: 78.6\%).

\begin{table}[t]
  \centering
  \small
  \caption{\textbf{Effect of image access on reader performance.} $\Delta = \mathrm{acc}_{+\mathrm{img}} - \mathrm{acc}_{-\mathrm{img}}$. Positive values mean the image helps. 95\% CIs and one-sided $p$-values from $10^4$-resample unpaired bootstrap of $\Delta$.}
  \label{tab:image-effect}
  \begin{tabular}{llrll}
    \toprule
    Reader & Question type & $\Delta$ [\%] & 95\% CI [\%] & $p$ \\
    \midrule
    $\mathrm{VP}_3$ & MCQ   & \deltaImgVPthreeMCQ   & $[\ciloImgVPthreeMCQ, \cihiImgVPthreeMCQ]$     & \pvalImgVPthreeMCQ \\
    $\mathrm{VP}_3$ & KPrim & \deltaImgVPthreeKPrim & $[\ciloImgVPthreeKPrim, \cihiImgVPthreeKPrim]$ & \pvalImgVPthreeKPrim \\
    $\mathrm{PP}_1$ & MCQ   & \deltaImgHPoneMCQ     & $[\ciloImgHPoneMCQ, \cihiImgHPoneMCQ]$         & \pvalImgHPoneMCQ \\
    $\mathrm{PP}_1$ & KPrim & \deltaImgHPoneKPrim   & $[\ciloImgHPoneKPrim, \cihiImgHPoneKPrim]$     & \pvalImgHPoneKPrim \\
    \bottomrule
  \end{tabular}
\end{table}

\subsection{Cross-benchmark generalization on PathMMU}

We also tested transfer in the reverse direction by evaluating $\tox$ alongside PathChat+ and GPT-5.4 on PathMMU, a reference benchmark for human pathology. PathMMU includes 33,428 MCQs paired with 24,067 pathology images. This setting probes whether specialization to non-human tissue leads to a collapse in performance when the model is transferred to human pathology. On PathMMU test set (\textbf{Appendix \autoref{tab:pathmmu}}), $\toxgemma$ reached 68.1\% [95\% CI 67.1--69.1], statistically similar to GPT-5.4 (68.8\% [67.8--69.8], $\Delta=-0.7$\% $[-1.8, +0.4]$, $P=0.20$, paired bootstrap, $10^4$ resamples). PathChat+ (70.2\% [69.2--71.2]) performs 2.1\% $[-3.3, -1.0]$ ($P<0.001$) better than $\toxgemma$. $\toxqwen$ reached 65.0\% [64.0--66.0], 3.1\% below its Gemma counterpart ($P<0.001$). These results show substantial cross-domain transfer, where veterinary specialization preserves competitive performance on human pathology, although it does not match a fully human-pathology-specialized model.

\section{Discussion}
\label{sec:discussion}

\paragraph{Summary.} VIPER addresses a missing piece in AI for pathology by providing an expert-curated benchmark for ROI-level vision-language evaluation in rat toxicologic pathology. Its design emphasizes visual grounding through expert-authored questions, human-in-the-loop refinement, and adversarial filtering. Our results show that human pathology-specialized models and general-purpose frontier models have limited transfer to veterinary pathology, while veterinary-specific instruction tuning yields consistent gains across backbones. VIPER therefore provides both a benchmark for measuring progress in non-human pathology VLMs and a template for building visually grounded biomedical evaluation datasets.

\paragraph{Limitations.} VIPER includes several limitations. First, VIPER is a ROI-level benchmark, which is only a step toward fully AI-assisted toxicologic pathology workflows that require slide-level review and dose-group comparison. \guillaume{This design reflects the fact that slide-level findings in toxicology studies are often coarse or incomplete, whereas ROIs enable more objective questions grounded in local morphology.} Second, although VIPER covers \guillaume{\norgans{} organ systems} commonly encountered in preclinical toxicology, it does not yet span the full organ spectrum required for preclinical safety assessment. \guillaume{The brain, the spinal cord and peripheral nerve are absent in particular, owing to a lack of available data. VIPER also does not cover the range of species used in toxicology studies, and multi-species extension is the natural next step.} Third, VIPER remains modest in size relative to broad public benchmarks, with 419 images and 1,251 questions. Therefore, organ-level and category-level analyses should be interpreted with appropriate caution, especially for smaller groups. \guillaume{Size, however, trades against the risk of contamination and data leakage. In VIPER, every image is a fresh ROI crop from a whole-slide image that has never been published as a standalone figure and has never been captioned, so no natural-language description of any VIPER field exists in a public corpus. Benchmarks assembled from educational content, PubMed figures, and social media draw on exactly the sources most likely to appear in a frontier model's pretraining data, so a smaller benchmark that a model cannot have memorized measures something a larger scraped one cannot.} Fourth, free-text scoring is inherently subjective and can introduce noise, even with structured rubrics and LLM-based judging.

\paragraph{Impact.} \guillaume{Toxicologic pathology is a key part of preclinical drug development. Findings from animal studies help determine whether a drug candidate can advance to testing in humans. VIPER provides a standard benchmark to evaluate vision-language models in this domain and extends current evaluation beyond human pathology. Our results show that strong performance in human pathology does not ensure strong performance in veterinary pathology. They also show that domain-specific training can improve performance across different model backbones. 
Performance on normal tissue provides a complementary test of model reliability. In safety assessment, models must be able to identify lesions accurately while avoiding false-positive findings in normal tissue. VIPER provides a basis for the development and evaluation of models that could support pathologists with screening, triage, and lesion characterization. Future work should extend this evaluation to more organs and species, whole-slide images, multiple animals and dose groups, and treatment-related findings. Because errors in preclinical safety assessment can affect drug-development decisions, these systems require careful evaluation under expert supervision before use in routine practice.}

\begin{ack}

\paragraph{Funding.} \guillaume{G.J. was funded by the European Union (ERC, DeepSPIM, 101219838). L.L.W, N.G.R., A.H.S., L.P.L., and F.M. were funded by Mass General Brigham (MGB). L.L.W. was funded by the U.S. National Science Foundation's Graduate Research Fellowship Program (NSF GRFP). Views and opinions expressed are however those of the author(s) only and do not necessarily reflect those of the European Union or the European Research Council. Neither the European Union nor the granting authority can be held responsible for them.}

\paragraph{Competing interests.} \guillaume{L.L.W., J.N.K., L.P.L., F.M., and G.J. hold equity in Tremont AI, a company developing AI solutions for preclinical toxicologic pathology.}
\end{ack}

\clearpage
\bibliographystyle{plain}
\bibliography{main}

\newpage
\input{checklist.tex}

\newpage 
\input{appendix.tex}

\end{document}

%% file: results/additional_vars.tex
\newcommand{\krippAMCQ}{0.875}
\newcommand{\krippAKPrim}{0.775}
\newcommand{\krippANmcq}{30}
\newcommand{\krippANkprim}{148}
\newcommand{\krippBMCQ}{0.789}
\newcommand{\krippBKPrim}{0.637}
\newcommand{\krippBNmcq}{29}
\newcommand{\krippBNkprim}{132}
\newcommand{\fleissAMCQ}{0.874}
\newcommand{\fleissAKPrim}{0.774}
\newcommand{\fleissBMCQ}{0.786}
\newcommand{\fleissBKPrim}{0.636}
\newcommand{\deltaImgVPthreeMCQ}{31.6}
\newcommand{\ciloImgVPthreeMCQ}{7.9}
\newcommand{\cihiImgVPthreeMCQ}{52.3}
\newcommand{\pvalImgVPthreeMCQ}{0.003}
\newcommand{\deltaImgVPthreeKPrim}{35.6}
\newcommand{\ciloImgVPthreeKPrim}{18.4}
\newcommand{\cihiImgVPthreeKPrim}{52.1}
\newcommand{\pvalImgVPthreeKPrim}{<0.001}
\newcommand{\deltaImgHPoneMCQ}{35.7}
\newcommand{\ciloImgHPoneMCQ}{11.8}
\newcommand{\cihiImgHPoneMCQ}{59.4}
\newcommand{\pvalImgHPoneMCQ}{0.003}
\newcommand{\deltaImgHPoneKPrim}{12.4}
\newcommand{\ciloImgHPoneKPrim}{-6.1}
\newcommand{\cihiImgHPoneKPrim}{30.3}
\newcommand{\pvalImgHPoneKPrim}{0.096}
\newcommand{\accVPtwoSetAMCQ}{93.3}
\newcommand{\accVPtwoSetAMCQcilo}{83.3}
\newcommand{\accVPtwoSetAMCQcihi}{100.0}
\newcommand{\accVPtwoSetAMCQn}{30}
\newcommand{\accVPtwoSetAKPrim}{75.7}
\newcommand{\accVPtwoSetAKPrimcilo}{63.5}
\newcommand{\accVPtwoSetAKPrimcihi}{86.5}
\newcommand{\accVPtwoSetAKPrimn}{37}
\newcommand{\accVPtwoSetBMCQ}{93.1}
\newcommand{\accVPtwoSetBMCQcilo}{82.8}
\newcommand{\accVPtwoSetBMCQcihi}{100.0}
\newcommand{\accVPtwoSetBMCQn}{29}
\newcommand{\accVPtwoSetBKPrim}{72.7}
\newcommand{\accVPtwoSetBKPrimcilo}{60.6}
\newcommand{\accVPtwoSetBKPrimcihi}{84.8}
\newcommand{\accVPtwoSetBKPrimn}{33}
\newcommand{\accVPthreeSetAMCQ}{83.3}
\newcommand{\accVPthreeSetAMCQcilo}{70.0}
\newcommand{\accVPthreeSetAMCQcihi}{96.7}
\newcommand{\accVPthreeSetAMCQn}{30}
\newcommand{\accVPthreeSetAKPrim}{81.1}
\newcommand{\accVPthreeSetAKPrimcilo}{70.3}
\newcommand{\accVPthreeSetAKPrimcihi}{90.5}
\newcommand{\accVPthreeSetAKPrimn}{37}
\newcommand{\accHPoneSetBMCQ}{72.4}
\newcommand{\accHPoneSetBMCQcilo}{55.2}
\newcommand{\accHPoneSetBMCQcihi}{86.2}
\newcommand{\accHPoneSetBMCQn}{29}
\newcommand{\accHPoneSetBKPrim}{65.2}
\newcommand{\accHPoneSetBKPrimcilo}{51.5}
\newcommand{\accHPoneSetBKPrimcihi}{78.8}
\newcommand{\accHPoneSetBKPrimn}{33}
\newcommand{\iccA}{0.658}
\newcommand{\iccAn}{33}
\newcommand{\iccApearson}{0.651}
\newcommand{\iccAmeanVPtwo}{76.8}
\newcommand{\iccAmeanVPthree}{75.5}
\newcommand{\iccB}{0.428}
\newcommand{\iccBn}{38}
\newcommand{\iccBpearson}{0.422}
\newcommand{\iccBmeanVPtwo}{49.3}
\newcommand{\iccBmeanHPone}{50.1}
\newcommand{\deltaVPvHPSetBMCQ}{20.7}
\newcommand{\ciloVPvHPSetBMCQ}{6.9}
\newcommand{\cihiVPvHPSetBMCQ}{37.9}
\newcommand{\pvalVPvHPSetBMCQ}{0.001}
\newcommand{\nVPvHPSetBMCQ}{29}
\newcommand{\meanVPvHPSetBMCQVP}{93.1}
\newcommand{\meanVPvHPSetBMCQHP}{72.4}
\newcommand{\deltaVPvHPSetBKPrim}{7.6}
\newcommand{\ciloVPvHPSetBKPrim}{-9.1}
\newcommand{\cihiVPvHPSetBKPrim}{24.2}
\newcommand{\pvalVPvHPSetBKPrim}{0.208}
\newcommand{\nVPvHPSetBKPrim}{33}
\newcommand{\meanVPvHPSetBKPrimVP}{72.7}
\newcommand{\meanVPvHPSetBKPrimHP}{65.2}
\newcommand{\deltaVPvHPSetBFT}{-0.8}
\newcommand{\ciloVPvHPSetBFT}{-14.0}
\newcommand{\cihiVPvHPSetBFT}{13.1}
\newcommand{\pvalVPvHPSetBFT}{0.553}
\newcommand{\nVPvHPSetBFT}{38}
\newcommand{\meanVPvHPSetBFTVP}{49.3}
\newcommand{\meanVPvHPSetBFTHP}{50.1}

\newcommand{\ftJudgeModel}{gpt-5.4}
\newcommand{\accVPtwoSetAFT}{76.8}
\newcommand{\accVPtwoSetAFTcilo}{66.0}
\newcommand{\accVPtwoSetAFTcihi}{86.5}
\newcommand{\accVPtwoSetAFTn}{33}
\newcommand{\accVPtwoSetBFT}{49.3}
\newcommand{\accVPtwoSetBFTcilo}{37.0}
\newcommand{\accVPtwoSetBFTcihi}{61.5}
\newcommand{\accVPtwoSetBFTn}{38}
\newcommand{\accVPthreeSetAFT}{75.5}
\newcommand{\accVPthreeSetAFTcilo}{64.5}
\newcommand{\accVPthreeSetAFTcihi}{85.3}
\newcommand{\accVPthreeSetAFTn}{33}
\newcommand{\accHPoneSetBFT}{50.1}
\newcommand{\accHPoneSetBFTcilo}{37.2}
\newcommand{\accHPoneSetBFTcihi}{62.9}
\newcommand{\accHPoneSetBFTn}{38}

%% file: results/main_results.tex
\begin{table}[t]
  \caption{VIPER benchmark results. Overall score and breakdown by question type. Best result per column in \textbf{bold}, second-best \underline{underlined}. $n=\nquestions{}$ questions. 95\% bootstrap confidence intervals (10{,}000 resamples). MCQ scores are mean across 5 cyclic-shift rotations of the answer ordering.}
  \label{tab:main-results}
  \centering
  \small
  \begin{tabular}{llcccc}
    \toprule
    Model & Domain & MCQ & KPrim & Free-Text & Overall \\
    \midrule
    $\toxqwen$ & Veterinary pathology & \textbf{67.1\,\tiny{[63.2,71.0]}} & \underline{61.8\,\tiny{[58.1,65.6]}} & \textbf{58.3\,\tiny{[54.4,62.4]}} & \textbf{62.4\,\tiny{[60.2,64.6]}} \\
    $\toxgemma$ & Veterinary pathology & \underline{65.2\,\tiny{[61.4,69.0]}} & \textbf{64.1\,\tiny{[60.4,67.9]}} & 54.3\,\tiny{[50.2,58.3]} & \underline{61.2\,\tiny{[59.0,63.5]}} \\
    GPT-5.4 & General-purpose & 58.5\,\tiny{[54.2,62.5]} & 54.3\,\tiny{[50.6,58.1]} & \underline{55.1\,\tiny{[50.8,59.3]}} & 56.0\,\tiny{[53.7,58.3]} \\
    Gemma 4 & General-purpose & 60.7\,\tiny{[56.7,64.7]} & 54.1\,\tiny{[50.2,57.9]} & 48.3\,\tiny{[44.0,52.6]} & 54.4\,\tiny{[52.0,56.7]} \\
    Qwen 3.5-27B & General-purpose & 60.0\,\tiny{[56.0,63.8]} & 46.6\,\tiny{[42.9,50.4]} & 50.6\,\tiny{[46.4,54.9]} & 52.4\,\tiny{[50.2,54.7]} \\
    PathChat+ & Human pathology & 58.7\,\tiny{[54.7,62.9]} & 41.5\,\tiny{[37.8,45.4]} & 52.7\,\tiny{[48.7,56.7]} & 51.0\,\tiny{[48.7,53.3]} \\
    Claude Sonnet 4.6 & General-purpose & 54.6\,\tiny{[50.6,58.6]} & 47.1\,\tiny{[43.4,50.7]} & 42.8\,\tiny{[38.8,46.9]} & 48.2\,\tiny{[45.9,50.5]} \\
    GPT-5.4-mini & General-purpose & 48.0\,\tiny{[43.8,52.1]} & 45.9\,\tiny{[42.1,49.8]} & 50.0\,\tiny{[45.9,54.2]} & 48.0\,\tiny{[45.7,50.4]} \\
    Gemini 2.5 Flash & General-purpose & 52.8\,\tiny{[48.8,56.8]} & 45.0\,\tiny{[41.4,48.8]} & 25.2\,\tiny{[21.5,29.0]} & 41.0\,\tiny{[38.8,43.2]} \\
    Patho-R1-7B & Human pathology & 46.1\,\tiny{[42.2,49.8]} & 16.3\,\tiny{[13.6,19.2]} & 46.0\,\tiny{[42.0,50.2]} & 36.1\,\tiny{[34.0,38.2]} \\
    Patho-R1-3B & Human pathology & 39.5\,\tiny{[36.2,42.8]} & 12.9\,\tiny{[10.6,15.3]} & 36.1\,\tiny{[32.3,40.1]} & 29.5\,\tiny{[27.7,31.5]} \\
    PathGen-LLaVA & Human pathology & 18.7\,\tiny{[16.1,21.2]} & 28.4\,\tiny{[25.1,31.9]} & 37.9\,\tiny{[33.7,42.2]} & 28.3\,\tiny{[26.3,30.3]} \\
    GPT-5.4-nano & General-purpose & 24.4\,\tiny{[21.6,27.3]} & 30.1\,\tiny{[26.9,33.3]} & 27.7\,\tiny{[24.2,31.4]} & 27.4\,\tiny{[25.5,29.3]} \\
    MedGemma-4B & Human pathology & 29.9\,\tiny{[27.1,32.9]} & 18.4\,\tiny{[15.6,21.3]} & 26.0\,\tiny{[22.6,29.6]} & 24.8\,\tiny{[23.0,26.6]} \\
    Quilt-LLaVA & Human pathology & 27.5\,\tiny{[24.9,30.4]} & 2.1\,\tiny{[1.2,3.0]} & 28.0\,\tiny{[24.7,31.6]} & 19.2\,\tiny{[17.7,20.8]} \\
    LLaVA-Med & Human pathology & 17.0\,\tiny{[15.8,18.3]} & 6.6\,\tiny{[5.0,8.5]} & 24.8\,\tiny{[21.4,28.2]} & 16.2\,\tiny{[14.8,17.5]} \\
    \bottomrule
  \end{tabular}
\end{table}

%% file: checklist.tex
\section*{NeurIPS Paper Checklist}

\begin{enumerate}

\item {\bf Claims}
    \item[] Question: Do the main claims made in the abstract and introduction accurately reflect the paper's contributions and scope?
    \item[] Answer: \answerYes{}.
    \item[] Justification: The Abstract and Introduction clearly state the contributions of VIPER, the first expert-curated benchmark for evaluating vision-language models in veterinary toxicologic pathology. The claims made in the Abstract and Introduction are limited to dataset construction, expert validation, visual grounding, and benchmarking of existing models, and are supported by the dataset statistics, curation protocol, reader study, and model evaluation results reported in the paper.

\item {\bf Limitations}
    \item[] Question: Does the paper discuss the limitations of the work performed by the authors?
    \item[] Answer: \answerYes{}.
    \item[] Justification: The paper discusses limitations of the study in the Discussion section, including the restricted organ and species scope, the focus on H\&E-stained rat tissue, the use of selected ROIs rather than whole-slide diagnostic workflows, and the fact that model performance on VIPER should not be interpreted as evidence of readiness for deployment in toxicologic pathology.

\item {\bf Theory assumptions and proofs}
    \item[] Question: For each theoretical result, does the paper provide the full set of assumptions and a complete (and correct) proof?
    \item[] Answer: \answerNA{}.
    \item[] Justification: The paper does not introduce theoretical results, theorems, or formal proofs. The contribution is a benchmark dataset, curation protocol, expert validation study, and empirical evaluation of vision-language models.

    \item {\bf Experimental result reproducibility}
    \item[] Question: Does the paper fully disclose all the information needed to reproduce the main experimental results of the paper to the extent that it affects the main claims and/or conclusions of the paper (regardless of whether the code and data are provided or not)?
    \item[] Answer: \answerYes{}.
    \item[] Justification: The paper describes the benchmark composition, question formats, expert curation procedure, visual-grounding checks, model evaluation protocol, prompting setup, scoring criteria, and statistical analysis used to produce the main results. These details provide a reasonable path to reproduce or verify the reported evaluations. In addition, we release the benchmark on HuggingFace and the code for running the benchmark on GitHub under the CC-BY-NC-4.0 license.

\item {\bf Open access to data and code}
    \item[] Question: Does the paper provide open access to the data and code, with sufficient instructions to faithfully reproduce the main experimental results, as described in supplemental material?
    \item[] Answer: \answerYes{}.
    \item[] Justification: VIPER is released to the research community as a new benchmark, with documentation and instructions for reproducing the main evaluation results. The release includes the benchmark questions and images (released on HuggingFace) as well as the evaluation code and scoring protocol (released on GitHub). The code includes clear instructions for installation and running the benchmark. In addition, we provide helpers to let users benchmark new models. A README file summarizes all steps. \guillaume{The $\tox$ model weights and the instruction-tuning corpus are not released under current institutional policy. the full training recipe is documented in \textbf{Appendix~\ref{app:toxscribe-training}}.}

\item {\bf Experimental setting/details}
    \item[] Question: Does the paper specify all the training and test details (e.g., data splits, hyperparameters, how they were chosen, type of optimizer) necessary to understand the results?
    \item[] Answer: \answerYes{}.
    \item[] Justification: The paper describes the benchmark composition, question formats, expert curation procedure, visual-grounding checks, model evaluation protocol, prompting setup, scoring criteria, and statistical analysis used to produce the main results. ToxScribe training data composition, optimizer (AdamW with cosine schedule), LoRA configuration (rank 64, $\alpha{=}128$, dropout 0.05), peak learning rate ($5{\times}10^{-5}$), warmup fraction, batch size, and total training steps are provided in the paper. These details provide a reasonable path to reproduce or verify the reported evaluations. In addition, we release the benchmark on HuggingFace and the code for running the benchmark on GitHub under the CC-BY-NC-4.0 license.

\item {\bf Experiment statistical significance}
    \item[] Question: Does the paper report error bars suitably and correctly defined or other appropriate information about the statistical significance of the experiments?
    \item[] Answer: \answerYes{}.
    \item[] Justification: The paper reports uncertainty and statistical significance for the key comparisons supporting the main claims, including expert agreement and the comparison between veterinary and human pathologist performance. Statistical procedures such as agreement metrics and paired bootstrap testing are described where used. All claims are backed by an appropriate and justified statistical test. 

\item {\bf Experiments compute resources}
    \item[] Question: For each experiment, does the paper provide sufficient information on the computer resources (type of compute workers, memory, time of execution) needed to reproduce the experiments?
    \item[] Answer: \answerYes{}.
    \item[] Justification: The paper clearly states that $\toxqwen$ was trained on 8$\times$ NVIDIA H200 GPUs for approximately 90 hours wall-clock (one epoch, 62{,}201 steps over 3.98M instruction pairs), and $\toxgemma$ for approximately 30 hours on the same hardware, for a combined ${\approx}$1{,}000 GPU-hours of training compute. Inference for both $\tox$ variants and other open-weight baselines was performed on 2$\times$ NVIDIA RTX Pro 6000 Blackwell workstation GPUs, requiring approximately 100 GPU-hours across the reported benchmark and image-ablation passes. Closed-weight commercial models (the GPT-5.4 family, Claude Sonnet~4.6, Gemini~2.5 Flash) were accessed via their respective APIs without local compute. Preliminary experiments, checkpoint sweeps, and exploratory training configurations added approximately 200 GPU-hours.

\item {\bf Code of ethics}
    \item[] Question: Does the research conducted in the paper conform, in every respect, with the NeurIPS Code of Ethics \url{https://neurips.cc/public/EthicsGuidelines}?
    \item[] Answer: \answerYes{}.
    \item[] Justification: The research uses preclinical animal pathology images and expert-authored benchmark questions. The paper describes the provenance of the data (and their license), the expert curation process, and the intended research use of the benchmark. The work is not presented as a deployable diagnostic system and does not involve clinical decision-making for human subjects.

\item {\bf Broader impacts}
    \item[] Question: Does the paper discuss both potential positive societal impacts and negative societal impacts of the work performed?
    \item[] Answer: \answerYes{}.
    \item[] Justification: The paper discusses positive impacts, including improved evaluation of vision-language models for toxicologic pathology, better detection of visually ungrounded model behavior, and support for safer AI development in preclinical drug safety. It also discusses risks, including overinterpretation of benchmark performance, inappropriate deployment without expert oversight, and possible misuse of model outputs in regulated scientific workflows.

\item {\bf Safeguards}
    \item[] Question: Does the paper describe safeguards that have been put in place for responsible release of data or models that have a high risk for misuse (e.g., pre-trained language models, image generators, or scraped datasets)?
    \item[] Answer: \answerYes{}.
    \item[] Justification: The paper releases a benchmark dataset and evaluation protocol. The benchmark is intended for research evaluation of pathology VLMs and not for deployment. The paper clear frames VIPER as an evaluation resource and not a clinical or regulatory decision system.

\item {\bf Licenses for existing assets}
    \item[] Question: Are the creators or original owners of assets (e.g., code, data, models), used in the paper, properly credited and are the license and terms of use explicitly mentioned and properly respected?
    \item[] Answer: \answerYes{}.
    \item[] Justification: The paper cites the original sources of the preclinical pathology data, existing benchmarks, baseline models, and software assets used in the study, and credits the model developers for all evaluated systems. \guillaume{Each image is redistributed under the license of its source repository, namely CC BY-SA 2.1 JP for Open TG-GATEs and CC BY-NC 4.0 for MMO.}

\item {\bf New assets}
    \item[] Question: Are new assets introduced in the paper well documented and is the documentation provided alongside the assets?
    \item[] Answer: \answerYes{}.
    \item[] Justification: The paper introduces VIPER as a new benchmark asset and documents its image sources, organ composition, question types, annotation and validation process, expert review procedure, intended use, \guillaume{licensing,} and limitations. \guillaume{Each image is released under the license of its source repository, and all other benchmark content under CC BY-NC 4.0.} The released materials are accompanied by documentation for evaluation and interpretation of the benchmark.

\item {\bf Crowdsourcing and research with human subjects}
    \item[] Question: For crowdsourcing experiments and research with human subjects, does the paper include the full text of instructions given to participants and screenshots, if applicable, as well as details about compensation (if any)? 
    \item[] Answer: \answerNA{}.
    \item[] Justification: The paper does not involve crowdsourcing. The reader study described in Section~\ref{sec:reader-study} was conducted with three board-certified veterinary pathologists and one board-certified physician pathologist participating as research collaborators (acknowledged in the author list); no crowd workers were employed and no monetary compensation was provided beyond standard research-collaboration arrangements.

\item {\bf Institutional review board (IRB) approvals or equivalent for research with human subjects}
    \item[] Question: Does the paper describe potential risks incurred by study participants, whether such risks were disclosed to the subjects, and whether Institutional Review Board (IRB) approvals (or an equivalent approval/review based on the requirements of your country or institution) were obtained?
    \item[] Answer: \answerNA{}.
    \item[] Justification: The paper does not involve human subjects or patient data. The benchmark is based on preclinical animal pathology images and expert-authored or expert-validated questions. Therefore, human-subject IRB approval is not applicable.

\item {\bf Declaration of LLM usage}
    \item[] Question: Does the paper describe the usage of LLMs if it is an important, original, or non-standard component of the core methods in this research? Note that if the LLM is used only for writing, editing, or formatting purposes and does \emph{not} impact the core methodology, scientific rigor, or originality of the research, declaration is not required.
    \item[] Answer: \answerYes{}.
    \item[] Justification: The paper describes the use of LLMs as part of the benchmark construction and evaluation workflow where relevant, including their role in generating or assisting with question variants under expert supervision and their use as evaluated VLM baselines. Expert review and validation are used to ensure that the final benchmark content is visually grounded and scientifically appropriate.

\end{enumerate}

%% file: appendix.tex
\appendix

\setcounter{figure}{0}
\setcounter{table}{0}

\renewcommand{\figurename}{Appendix Figure}
\renewcommand{\tablename}{Appendix Table}

\renewcommand{\thefigure}{\arabic{figure}}
\renewcommand{\thetable}{\arabic{table}}

\section{Additional information about VIPER}

\subsection{Training details of $\tox$.}
\label{app:toxscribe-training}

$\toxqwen$ and $\toxgemma$ are LoRA fine-tuned versions of \texttt{Qwen/Qwen3.5-27B} and \texttt{google/gemma-4-31B-it}, respectively, sharing the same instruction-tuning corpus.

\paragraph{Corpus assembly.} Training data were assembled from permissively reusable pathology sources, including PubMed Open Access PDFs, regulatory agency documents, and teaching materials. PDF documents were parsed into markdown using Docling~\cite{Docling}. Images smaller than 384$\times$384 pixels were discarded. We then trained an H\&E image detector to retain relevant histopathology content and used a YOLOv11-based object detector~\cite{yolo11_ultralytics} to decompose multi-panel figures into sub-images. Extracted subpanels were associated with their corresponding captions or sub-captions using GPT-5-mini. The final corpus contained approximately 345{,}000 pathology ROIs and 3.98 million instruction pairs derived from image-caption pairs. The corpus spans all major organ systems across multiple species and contains a large proportion of rodent tissue.

\paragraph{Optimization.} Both $\toxqwen$ and $\toxgemma$ were optimized with AdamW using a cosine learning-rate schedule, $\beta_1{=}0.9$, and $\beta_2{=}0.999$. LoRA adapters were applied to all attention and MLP projections of the language model, with rank 64, $\alpha{=}128$, and dropout 0.05, while the vision backbone was kept frozen. Images were sampled across magnifications, from low-resolution thumbnails to high-resolution histology crops. Training was run for 62{,}201 steps, corresponding to one full epoch over the 3.98 million instruction pairs, with a peak learning rate of $5\times 10^{-5}$, 3\% warmup, and an effective batch size of 64, using per-device batch size 2 and 4 gradient-accumulation steps on 8$\times$ H200 GPUs.

\paragraph{Inference.} At inference time, $\toxqwen$ and $\toxgemma$ were served with vLLM at temperature $0.0$. Image inputs were passed in-line as base64-encoded PNGs alongside the question prompt under each backbone's native chat template. Inference was run on 2$\times$ RTX Pro 6000 Blackwell workstation GPUs.

\subsection{VIPER distribution by organ system and question category}

\input{taxonomy/anatomic_structures}

\input{taxonomy/appendix_taxonomy}

\clearpage

\subsection{Reader study analysis}

\input{results/agreement_table.tex}

\input{reader_study/pairwise_agreement}

\input{reader_study/free_text_aggreement}

\clearpage

\subsection{Free-text LLM judge robustness}
\label{app:judge-robustness}

We compute free-text accuracy with a single LLM judge (Sec.~\ref{sec:eval-protocol}), which raises two concerns: (i) that scoring is sensitive to stochastic variation in the judge, and (ii) that scores reflect properties of the chosen model family instead of the model under evaluation. To address both, we re-score the free-text predictions of $\toxqwen$, GPT-5.4, and PathChat+ on VIPER under (i)~a second, independent call to the same GPT-5.4 judge, to bound within-judge stochasticity, and (ii)~Claude Opus 4.7 from an independent vendor (Anthropic), to test for cross-vendor robustness. Both re-scoring conditions use the same prompt and composite formula as the main pipeline (\textbf{\autoref{tab:judge-crossmodel}}).

\input{results/judge_robustness}

Pooling across the three models in \textbf{\autoref{tab:judge-crossmodel}} ($n=3\times\nfreetext{}=1{,}254$ paired predictions), the GPT-5.4 judge and Claude Opus 4.7 produce nearly collinear item-level scores: Pearson $r = 0.98$ and Spearman $\rho = 0.95$. The two judges therefore agree closely on which responses are good or poor, and their disagreement is concentrated in the absolute level, not the item-level ranking. A second, independent call to the same GPT-5.4 judge is even closer to the headline pipeline (Pearson $r = 0.99$) and shifts each model's mean composite by only $0.8$--$1.1$~\% (paired-bootstrap CI on $\toxqwen$: $[+0.28, +1.41]$), so within-judge stochasticity is small. The $\toxqwen$ advantage over PathChat+ on free text is $5.4$--$5.7$~\% under both the replicate GPT-5.4 judge and Claude Opus 4.7, and is statistically significant under each (one-sided paired bootstrap $p \le 0.010$, $10^4$ item-level resamples; Holm-adjusted $p \le 0.008$ over the two-judge family). The $\toxqwen$ advantage over GPT-5.4 (the model) on free text is $1.6$--$3.2$~\% but does not reach significance under either judge; this is consistent with \textbf{\autoref{tab:main-results}}, in which free text is the closest of the three formats between these two models and the overall lead of $\toxqwen$ over GPT-5.4 is carried by MCQ and KPrim.

Absolute free-text scores are therefore calibrated to the chosen judge, while the directional free-text ranking between $\toxqwen$ and the strongest human-pathology-specialized baseline is stable across LLM judges from two independent vendors.

\clearpage

\subsection{Statistical comparison across lead models}

\input{results/category_significance.tex}

\subsection{Per-category performance}

\input{results/category_performance.tex}

\clearpage

\subsection{\guillaume{Performance by organ system}}

\input{results/per_organ_performance}

\subsection{Statistical comparison veterinary vs physician pathologist}

\input{reader_study/vet_vs_physician}

\clearpage

\subsection{Image-ablation matrix across question types}

\input{results/no_image_ablation}

\subsection{Cross-benchmark generalization on PathMMU}

\input{results/pathmmu}

\clearpage

\input{prompts}

\clearpage

%% file: taxonomy/anatomic_structures.tex
\begin{table}[h]
  \caption{\textbf{VIPER composition by organ system.} VIPER covers 23 distinct named anatomic structures
  drawn from seven organ systems.}
  \label{tab:structures}
  \centering
  \footnotesize
  \setlength{\tabcolsep}{3pt}
  \begin{tabular}{@{}p{2.3cm}p{5.0cm}ccccc@{}}
    \toprule
    Organ system & Named structures & Images & MCQ & KPrim & Free-Text & Total \\
    \midrule
    Urinary & kidney, urinary bladder, ureter, urethra & 137 & 137 & 135 & 136 & 408 \\
    \addlinespace
    Hepatobiliary & liver & 86 & 86 & 86 & 86 & 258 \\
    \addlinespace
    Endocrine & thyroid & 56 & 56 & 56 & 56 & 168 \\
    \addlinespace
    \makecell[tl]{Male reproductive\\ and sex glands}
      & prostate, ampullary gland, seminal vesicle, epididymis, ejaculatory duct, coagulating gland,
        periurethral gland, testis & 50 & 50 & 49 & 50 & 149 \\
    \addlinespace
    Digestive & salivary gland, stomach, colon, duodenum, small intestine, large intestine, pancreas
      & 40 & 40 & 38 & 40 & 118 \\
    \addlinespace
    Respiratory & lung & 28 & 28 & 28 & 28 & 84 \\
    \addlinespace
    Cardiovascular & heart & 22 & 22 & 22 & 22 & 66 \\
    \midrule
    \textbf{Total} & \textbf{23 structures} & \textbf{\nimages} & \textbf{\nmcq} & \textbf{\nkprim}
      & \textbf{\nfreetext} & \textbf{\nquestions} \\
    \bottomrule
  \end{tabular}
\end{table}

%% file: taxonomy/appendix_taxonomy.tex
\begin{table}[h]
  \small
  \centering
  \caption{\textbf{VIPER question categories.} $n$: number of questions per category. Definitions and prototype examples are drawn from the seed annotations.}
  \label{tab:taxonomy-definitions}
  \setlength{\tabcolsep}{4pt}
  \begin{tabular}{@{}p{2cm}rp{5.2cm}p{5.2cm}@{}}
    \toprule
    Category & $n$ & Definition & Prototype example \\
    \midrule
    \makecell[tl]{Anatomy \\ Identification} & 362
      & Name, verify, or identify a normal anatomic structure, tissue type, region, or cell type. Includes true/false verification of anatomy claims (either polarity) and ``unremarkable $X$ is represented'' seeds.
      & \emph{Q}: ``Which renal region is represented?'' \newline \emph{A}: Cortex \\[1ex]
    \makecell[tl]{Over-Reading \\ Probe} & 240
      & Hallucination-resistance probe. Two routes: true/false ``$X$ is within normal limits / changes are absent'' (answer true), or a specific false disease claim (answer false). Excludes false-anatomy claims.
      & \emph{Q}: ``Mild myositis is present.'' \newline \emph{A}: false \\[1ex]
    \makecell[tl]{Spatial \\ Localization} & 227
      & ``Where in the image is $X$?'' where the answer is a spatial region (quadrant, corner, edge, zone). Target may be anatomy or a lesion; output shape defines the category.
      & \emph{Q}: ``Where is tubular necrosis most evident?'' \newline \emph{A}: Top third and subcapsular along the right edge \\[1ex]
    \makecell[tl]{Pathology \\ Identification} & 221
      & Recognise that a specific pathology is present: open-ended naming of a real diagnosis, true/false verification of a lesion (answer true), or interpret-region-of-interest yielding a pathologic entity.
      & \emph{Q}: ``Define the main histopathologic change.'' \newline \emph{A}: Suppurative-necrotising hepatitis, focal \\[1ex]
    \makecell[tl]{Feature \\ Characterization} & 78
      & Describe attributes, distribution, grade, severity, or functional state of a named feature, lesion, or organ. Includes organ functional states (thyroid activity, bladder distension).
      & \emph{Q}: ``Add two descriptors for the hepatocellular necrosis.'' \newline \emph{A}: Coagulative, acute, focal \\[1ex]
    \makecell[tl]{Artifact \\ Recognition} & 63
      & Technical, processing, scanning, or slide-quality issues: artefacts, tissue absence, coverslip edges, autolysis, slide completeness.
      & \emph{Q}: ``Name the artefactual change.'' \newline \emph{A}: Stain precipitate \\[1ex]
    \makecell[tl]{Feature \\ Quantification} & 60
      & Numeric count, proportion, ratio, or extent of visible elements (normal or pathologic). Output has a numeric component.
      & \emph{Q}: ``Which proportion of the gland is affected by atrophy?'' \newline \emph{A}: Approx. 50--60\% \\
    \midrule
    \textbf{Total} & \textbf{1{,}251} & & \\
    \bottomrule
  \end{tabular}
\end{table}

%% file: results/agreement_table.tex
\begin{table}[h]
\centering
\caption{\textbf{Pairwise inter-annotator agreement.} MCQ uses choice-level agreement (5 categories, $p_e=1/5$). KPrim uses per-statement T/F agreement ($p_e=1/2$). VP1 is the benchmark author treated as a rater who always picks the correct choice. VP3 $\times$ HP1 has no overlapping questions.}
\small
\begin{tabular}{llrrrrrr}
\toprule
 & & \multicolumn{3}{c}{MCQ (choice-level)} & \multicolumn{3}{c}{KPrim (per-statement)} \\
\cmidrule(lr){3-5}\cmidrule(lr){6-8}
Pair & Set & $n$ & Agr. & $\kappa$ & $n$ & Agr. & $\kappa$ \\
\midrule
VP1 $\times$ VP2 & A & 30 & 0.933 & 0.917 & 148 & 0.858 & 0.716 \\
VP1 $\times$ VP2 & B & 29 & 0.931 & 0.914 & 132 & 0.848 & 0.697 \\
VP1 $\times$ VP3 & A & 30 & 0.833 & 0.792 & 148 & 0.905 & 0.811 \\
VP1 $\times$ HP1 & B & 29 & 0.724 & 0.655 & 132 & 0.818 & 0.636 \\
VP2 $\times$ VP3 & A & 30 & 0.867 & 0.833 & 148 & 0.899 & 0.797 \\
VP2 $\times$ HP1 & B & 29 & 0.724 & 0.655 & 132 & 0.788 & 0.576 \\
\bottomrule
\end{tabular}
\label{tab:interrater}
\end{table}

%% file: reader_study/pairwise_agreement.tex
\begin{table}[h]
  \centering
  \small
  \caption{\textbf{Summary of pairwise agreements among pathologists.} Three-rater reliability per set, summarizing all pairwise agreements in a single number. Krippendorff's $\alpha$ and Fleiss's $\kappa$ use different chance-correction conventions but produce near-identical values on this data. MCQ is analyzed at the choice level (5 categories); KPrim is analyzed per statement (2 categories). Set~A raters are $\mathrm{VP}_1, \mathrm{VP}_2, \mathrm{VP}_3$; Set~B raters are $\mathrm{VP}_1, \mathrm{VP}_2, \mathrm{PP}_1$.}
  \label{tab:multi-rater-reliability}
  \begin{tabular}{llrrr}
    \toprule
    Set & Question type & $n$ & Krippendorff's $\alpha$ & Fleiss's $\kappa$ \\
    \midrule
    A (3 vet.\ pathologists) & MCQ   & \krippANmcq   & \krippAMCQ   & \fleissAMCQ \\
    A (3 vet.\ pathologists) & KPrim & \krippANkprim & \krippAKPrim & \fleissAKPrim \\
    \midrule
    B (2 vet.\ + 1 physician pathologist) & MCQ   & \krippBNmcq   & \krippBMCQ   & \fleissBMCQ \\
    B (2 vet.\ + 1 physician pathologist) & KPrim & \krippBNkprim & \krippBKPrim & \fleissBKPrim \\
    \bottomrule
  \end{tabular}
\end{table}

%% file: reader_study/free_text_aggreement.tex
\begin{table}[h]
  \centering
  \small
  \caption{\textbf{Free-text inter-rater reliability on the two sets where two readers overlap.} Each response is scored on $[0,100]$ by a frozen two-axis LLM judge (\texttt{\ftJudgeModel{}}) with the per-question scoring rubric included in the prompt. Mean scores per rater are shown in percent; reliability is reported as ICC(2,1), the absolute-agreement single-measurement intraclass correlation, which is the continuous-score analogue of Cohen's $\kappa$ and the standard choice when the scoring unit is a continuous number rather than a discrete category. Pearson $r$ is shown for reference; it measures linear association but ignores absolute level. ICC and Pearson $r$ are unitless and reported as decimals.}
  \label{tab:ft-icc}
  \begin{tabular}{llrrrrr}
    \toprule
    Pair & Set & $n$ & Mean score (a) & Mean score (b) & Pearson $r$ & ICC(2,1) \\
    \midrule
    $\mathrm{VP}_2 \times \mathrm{VP}_3$ & A & \iccAn & \iccAmeanVPtwo & \iccAmeanVPthree & \iccApearson & \iccA \\
    $\mathrm{VP}_2 \times \mathrm{PP}_1$ & B & \iccBn & \iccBmeanVPtwo & \iccBmeanHPone   & \iccBpearson & \iccB \\
    \bottomrule
  \end{tabular}
\end{table}

%% file: results/judge_robustness.tex
\begin{table}[h]
  \centering
  \small
  \caption{\textbf{Free-text composite scores under independent LLM judges.} Each cell is the mean composite score over the same $\nfreetext{}$ free-text predictions, scored under the same prompt and composite formula as in \textbf{\autoref{tab:main-results}}. Values are in percent with 95\% percentile bootstrap CIs ($10^4$ item-level resamples). The first column reproduces the headline GPT-5.4 judge values from \textbf{\autoref{tab:main-results}}; the second column is a replicate scoring with the same GPT-5.4 judge to bound within-judge stochasticity; the third column is an independent judge from a different vendor (Anthropic).}
  \label{tab:judge-crossmodel}
  \setlength{\tabcolsep}{6pt}
  \begin{tabular}{lccc}
    \toprule
    Model & GPT-5.4 & GPT-5.4 (replicate) & Claude Opus 4.7 \\
    \midrule
    $\toxqwen$       & $58.3~[54.3, 62.2]$ & $59.1~[55.2, 63.2]$ & $56.0~[52.1, 59.8]$ \\
    GPT-5.4          & $55.0~[50.8, 59.3]$ & $56.0~[51.8, 60.2]$ & $54.4~[50.2, 58.5]$ \\
    PathChat+        & $52.7~[48.7, 56.7]$ & $53.8~[49.8, 57.6]$ & $50.3~[46.5, 54.1]$ \\
    \bottomrule
  \end{tabular}
\end{table}

%% file: results/category_significance.tex
\begin{table}[h]
  \caption{\textbf{Pair-wise comparison between lead models.} Paired item-level bootstrap significance tests across model-category leaders on the VIPER benchmark ($n=\nquestions{}$ questions, 10{,}000 resamples, one-sided $p$-value for $\Delta\le 0$). Category leaders are the highest-scoring model in each group. $\Delta$ is defined as the difference in accuracies: acc(Model A) - acc(Model B).}
  \label{tab:significance_categories}
  \centering
  \begin{tabular}{p{3.8cm}llcc}
    \toprule
    Comparison type & Model A & Model B & $\Delta$ (\%) [95\% CI] & $p$-value \\
    \midrule
    \makecell[l]{Best vet pathology\\vs.\ best human pathology} & ToxScribe (Qwen3.5) & PathChat+ & +11.5 [+9.0, +13.9] & $<$0.001 \\
    \makecell[l]{Best vet pathology\\vs.\ best frontier model} & ToxScribe (Qwen3.5) & GPT-5.4 & +6.5 [+4.0, +9.0] & $<$0.001 \\
    \makecell[l]{Best frontier model\\vs.\ best human pathology} & GPT-5.4 & PathChat+ & +5.0 [+2.4, +7.6] & $<$0.001 \\
    \bottomrule
  \end{tabular}
\end{table}

%% file: results/category_performance.tex
\begin{table}[h]
  \small
  \centering
  \caption{\textbf{VIPER mean performance (\%) by question category.} 95\% bootstrap confidence intervals (2{,}000 item-level resamples). Best mean per column in bold. $n$ is the number of questions assigned to that category on the frozen $\nquestions{}$-question set. Cell means are item-level averages within each (model, category); they are not directly comparable to the leaderboard overall scores (mean-of-means across question types).}
  \label{tab:category_performance}
  \resizebox{\linewidth}{!}{%
  \begin{tabular}{lccccccc}
    \toprule
     & \makecell{Anatomy \\ identification} & \makecell{Over-reading \\ probe} & \makecell{Spatial \\ localization} & \makecell{Pathology \\ identification} & \makecell{Feature \\ characterization} & \makecell{Artifact \\ recognition} & \makecell{Feature \\ quantification} \\
    $n$ & 362 & 240 & 227 & 221 & 78 & 63 & 60 \\
    \midrule
    ToxScribe (Qwen) & \makecell{\textbf{60.9} \\ {\scriptsize [55.8, 65.9]}} & \makecell{77.2 \\ {\scriptsize [71.6, 82.5]}} & \makecell{\textbf{54.0} \\ {\scriptsize [47.0, 60.5]}} & \makecell{\textbf{50.3} \\ {\scriptsize [44.0, 56.5]}} & \makecell{\textbf{62.9} \\ {\scriptsize [53.1, 72.4]}} & \makecell{57.0 \\ {\scriptsize [44.9, 68.8]}} & \makecell{45.1 \\ {\scriptsize [32.1, 57.5]}} \\
    ToxScribe (Gemma) & \makecell{57.0 \\ {\scriptsize [51.4, 62.4]}} & \makecell{\textbf{80.5} \\ {\scriptsize [75.7, 85.5]}} & \makecell{50.3 \\ {\scriptsize [44.1, 56.6]}} & \makecell{48.3 \\ {\scriptsize [42.1, 54.2]}} & \makecell{61.0 \\ {\scriptsize [50.4, 71.6]}} & \makecell{\textbf{62.7} \\ {\scriptsize [50.2, 75.0]}} & \makecell{\textbf{54.6} \\ {\scriptsize [41.4, 67.9]}} \\
    GPT-5.4 & \makecell{60.4 \\ {\scriptsize [55.2, 65.5]}} & \makecell{55.1 \\ {\scriptsize [48.1, 61.5]}} & \makecell{51.4 \\ {\scriptsize [44.4, 58.2]}} & \makecell{49.2 \\ {\scriptsize [42.8, 55.8]}} & \makecell{61.0 \\ {\scriptsize [49.9, 71.8]}} & \makecell{55.4 \\ {\scriptsize [44.3, 66.1]}} & \makecell{42.9 \\ {\scriptsize [29.9, 55.9]}} \\
    Gemma 4 & \makecell{50.0 \\ {\scriptsize [44.8, 55.3]}} & \makecell{60.1 \\ {\scriptsize [53.5, 66.5]}} & \makecell{52.3 \\ {\scriptsize [45.5, 59.4]}} & \makecell{44.5 \\ {\scriptsize [37.8, 51.0]}} & \makecell{44.9 \\ {\scriptsize [32.9, 56.5]}} & \makecell{57.2 \\ {\scriptsize [44.8, 68.8]}} & \makecell{44.0 \\ {\scriptsize [31.4, 56.8]}} \\
    Qwen 3.5-27B & \makecell{50.5 \\ {\scriptsize [45.1, 55.8]}} & \makecell{58.8 \\ {\scriptsize [52.3, 65.4]}} & \makecell{42.5 \\ {\scriptsize [35.8, 48.7]}} & \makecell{43.6 \\ {\scriptsize [37.2, 50.5]}} & \makecell{50.1 \\ {\scriptsize [40.2, 60.3]}} & \makecell{46.0 \\ {\scriptsize [33.4, 59.8]}} & \makecell{39.5 \\ {\scriptsize [27.0, 52.3]}} \\
    GPT-5.4-mini & \makecell{49.6 \\ {\scriptsize [44.2, 54.9]}} & \makecell{44.2 \\ {\scriptsize [37.3, 51.2]}} & \makecell{51.4 \\ {\scriptsize [44.2, 57.8]}} & \makecell{48.6 \\ {\scriptsize [42.3, 54.8]}} & \makecell{47.0 \\ {\scriptsize [37.1, 57.1]}} & \makecell{46.4 \\ {\scriptsize [33.9, 58.5]}} & \makecell{40.8 \\ {\scriptsize [29.3, 52.6]}} \\
    PathChat+ & \makecell{49.5 \\ {\scriptsize [44.3, 54.4]}} & \makecell{62.1 \\ {\scriptsize [55.5, 68.5]}} & \makecell{38.7 \\ {\scriptsize [31.9, 45.1]}} & \makecell{43.5 \\ {\scriptsize [37.5, 49.3]}} & \makecell{45.1 \\ {\scriptsize [35.1, 54.7]}} & \makecell{36.4 \\ {\scriptsize [24.6, 48.7]}} & \makecell{32.0 \\ {\scriptsize [20.8, 43.6]}} \\
    Claude Sonnet 4 & \makecell{46.4 \\ {\scriptsize [41.3, 51.5]}} & \makecell{54.3 \\ {\scriptsize [47.6, 60.9]}} & \makecell{39.0 \\ {\scriptsize [33.1, 45.0]}} & \makecell{40.2 \\ {\scriptsize [33.4, 46.4]}} & \makecell{39.0 \\ {\scriptsize [28.6, 48.9]}} & \makecell{57.5 \\ {\scriptsize [44.2, 70.4]}} & \makecell{33.0 \\ {\scriptsize [23.9, 42.1]}} \\
    Gemini 2.5 Flash & \makecell{39.8 \\ {\scriptsize [34.7, 44.9]}} & \makecell{33.8 \\ {\scriptsize [27.6, 40.2]}} & \makecell{43.5 \\ {\scriptsize [36.6, 50.3]}} & \makecell{26.5 \\ {\scriptsize [20.6, 32.7]}} & \makecell{32.7 \\ {\scriptsize [22.5, 43.1]}} & \makecell{35.0 \\ {\scriptsize [23.1, 47.2]}} & \makecell{14.6 \\ {\scriptsize [8.1, 21.5]}} \\
    PathGen-LLaVA & \makecell{33.3 \\ {\scriptsize [28.4, 38.1]}} & \makecell{53.4 \\ {\scriptsize [46.9, 60.0]}} & \makecell{23.8 \\ {\scriptsize [18.4, 30.0]}} & \makecell{25.0 \\ {\scriptsize [19.5, 31.0]}} & \makecell{39.2 \\ {\scriptsize [28.3, 50.2]}} & \makecell{19.6 \\ {\scriptsize [11.3, 30.0]}} & \makecell{23.0 \\ {\scriptsize [12.6, 33.6]}} \\
    Patho-R1-7B & \makecell{33.2 \\ {\scriptsize [28.2, 38.2]}} & \makecell{49.8 \\ {\scriptsize [42.8, 56.8]}} & \makecell{23.5 \\ {\scriptsize [18.2, 28.9]}} & \makecell{22.8 \\ {\scriptsize [17.4, 28.4]}} & \makecell{27.8 \\ {\scriptsize [18.6, 37.4]}} & \makecell{17.3 \\ {\scriptsize [8.9, 26.7]}} & \makecell{24.2 \\ {\scriptsize [13.6, 34.7]}} \\
    GPT-5.4-nano & \makecell{29.4 \\ {\scriptsize [25.4, 34.0]}} & \makecell{24.3 \\ {\scriptsize [18.9, 30.2]}} & \makecell{27.4 \\ {\scriptsize [22.1, 32.9]}} & \makecell{27.3 \\ {\scriptsize [22.2, 32.8]}} & \makecell{39.3 \\ {\scriptsize [29.2, 49.7]}} & \makecell{40.7 \\ {\scriptsize [29.0, 51.8]}} & \makecell{29.5 \\ {\scriptsize [18.7, 40.6]}} \\
    Patho-R1-3B & \makecell{26.3 \\ {\scriptsize [21.9, 30.7]}} & \makecell{42.0 \\ {\scriptsize [35.2, 48.9]}} & \makecell{16.3 \\ {\scriptsize [12.0, 20.8]}} & \makecell{16.6 \\ {\scriptsize [12.0, 21.3]}} & \makecell{19.3 \\ {\scriptsize [12.3, 27.6]}} & \makecell{13.9 \\ {\scriptsize [7.0, 22.7]}} & \makecell{22.7 \\ {\scriptsize [13.1, 33.1]}} \\
    MedGemma-4B & \makecell{24.7 \\ {\scriptsize [20.5, 29.0]}} & \makecell{28.8 \\ {\scriptsize [22.6, 35.1]}} & \makecell{20.1 \\ {\scriptsize [15.3, 25.0]}} & \makecell{15.0 \\ {\scriptsize [10.5, 19.5]}} & \makecell{24.6 \\ {\scriptsize [17.3, 32.3]}} & \makecell{9.2 \\ {\scriptsize [2.5, 17.1]}} & \makecell{25.0 \\ {\scriptsize [15.2, 34.9]}} \\
    LLaVA-Med & \makecell{18.6 \\ {\scriptsize [14.5, 23.0]}} & \makecell{17.4 \\ {\scriptsize [12.3, 23.0]}} & \makecell{17.6 \\ {\scriptsize [13.3, 22.4]}} & \makecell{10.3 \\ {\scriptsize [6.6, 14.1]}} & \makecell{11.2 \\ {\scriptsize [6.4, 16.3]}} & \makecell{6.1 \\ {\scriptsize [1.4, 12.7]}} & \makecell{21.3 \\ {\scriptsize [11.8, 31.3]}} \\
    Quilt-LLaVA & \makecell{14.1 \\ {\scriptsize [10.5, 18.0]}} & \makecell{31.0 \\ {\scriptsize [24.4, 37.6]}} & \makecell{9.9 \\ {\scriptsize [6.4, 13.7]}} & \makecell{7.1 \\ {\scriptsize [4.3, 10.3]}} & \makecell{14.4 \\ {\scriptsize [8.5, 20.7]}} & \makecell{7.6 \\ {\scriptsize [3.0, 13.0]}} & \makecell{15.5 \\ {\scriptsize [8.2, 24.0]}} \\
    \bottomrule
  \end{tabular}%
  }
\end{table}

%% file: results/per_organ_performance.tex
\begin{table}[h]
  \centering
  \footnotesize
  \setlength{\tabcolsep}{3pt}
  \caption{\textbf{Performance by organ system.} Overall score (\%) per (model, organ system) on the canonical $n=\nquestions{}$ bench, grouped into the \norgans{} organ systems of \textbf{Appendix~\autoref{tab:structures}}. MCQ scores are the mean across 5 cyclic-shift rotations of the answer ordering. Best per column in \textbf{bold}, second-best \underline{underlined}. Column headers list the organ system and its question count $n$.}
  \label{tab:per-organ}
  \begin{tabular}{lccccccc}
    \toprule
    Model & \makecell{Urinary\\($n$=408)} & \makecell{Hepatobiliary\\($n$=258)} & \makecell{Endocrine\\($n$=168)} & \makecell{Male\\Repro.\\($n$=149)} & \makecell{Digestive\\($n$=118)} & \makecell{Respiratory\\($n$=84)} & \makecell{Cardiovascular\\($n$=66)} \\
    \midrule
    ToxScribe (Qwen3.5) & \underline{65.0} & \textbf{64.3} & \underline{67.0} & \textbf{43.7} & \textbf{62.1} & \underline{65.9} & 66.0 \\
    ToxScribe (Gemma 4) & \textbf{65.3} & \underline{62.0} & \textbf{69.5} & 38.9 & 54.1 & \textbf{67.1} & 66.9 \\
    GPT-5.4 & 53.9 & 59.2 & 60.4 & \underline{40.4} & \underline{60.1} & 58.9 & \textbf{68.6} \\
    Gemma 4 & 54.5 & 55.6 & 63.9 & 35.0 & 50.0 & 61.1 & \underline{67.3} \\
    Qwen 3.5-27B & 54.4 & 55.1 & 53.4 & 33.5 & 51.9 & 59.6 & 61.4 \\
    PathChat+ & 55.8 & 50.2 & 54.6 & 35.1 & 52.6 & 45.0 & 55.9 \\
    Claude Sonnet 4.6 & 49.4 & 49.0 & 53.4 & 28.6 & 52.8 & 45.9 & 62.9 \\
    GPT-5.4-mini & 46.2 & 50.0 & 49.6 & 30.9 & 53.5 & 58.2 & 63.0 \\
    Gemini 2.5 Flash & 38.0 & 37.8 & 45.2 & 31.8 & 48.3 & 49.6 & 57.8 \\
    Patho-R1-7B & 39.1 & 37.4 & 35.0 & 26.9 & 35.2 & 33.1 & 43.3 \\
    Patho-R1-3B & 34.9 & 28.9 & 30.7 & 19.5 & 25.4 & 27.6 & 28.8 \\
    PathGen-LLaVA & 33.6 & 31.2 & 32.5 & 12.9 & 23.9 & 18.6 & 29.0 \\
    GPT-5.4-nano & 28.4 & 31.8 & 28.2 & 23.7 & 20.6 & 23.2 & 27.5 \\
    MedGemma-4B & 26.7 & 25.9 & 29.2 & 15.5 & 19.3 & 26.5 & 25.8 \\
    Quilt-LLaVA & 21.8 & 20.2 & 18.4 & 13.1 & 17.3 & 15.2 & 24.9 \\
    LLaVA-Med & 17.3 & 15.5 & 14.8 & 15.6 & 15.5 & 12.9 & 22.5 \\
    \bottomrule
  \end{tabular}
\end{table}

%% file: reader_study/vet_vs_physician.tex
\begin{table}[h]
  \centering
  \small
  \caption{\textbf{Comparison between $\mathrm{VP}_2$ and $\mathrm{PP}_1$ on Set~B.} Paired vet-versus-physician test on Set~B (both $\mathrm{VP}_2$ and $\mathrm{PP}_1$ answered the same items with image visible). Per-rater scores and $\Delta = \mathrm{acc}(\mathrm{VP}_2) - \mathrm{acc}(\mathrm{PP}_1)$ are reported in percent; positive $\Delta$ means the veterinary pathologist outperforms the physician pathologist. 95\% CIs and $p$-values are from $10^4$-resample paired bootstrap; $p$ is the one-sided probability that $\Delta \le 0$. Only one physician reader ($\mathrm{PP}_1$) participated, which limits the generalizability of the KPrim and free-text null results.}
  \label{tab:vet-vs-human}
  \begin{tabular}{lrrrll}
    \toprule
    Question type & $n$ & $\mathrm{VP}_2$ score & $\mathrm{PP}_1$ score & $\Delta$ [95\% CI] & $p$ \\
    \midrule
    MCQ       & \nVPvHPSetBMCQ   & \meanVPvHPSetBMCQVP   & \meanVPvHPSetBMCQHP   & \deltaVPvHPSetBMCQ~$[\ciloVPvHPSetBMCQ, \cihiVPvHPSetBMCQ]$     & \pvalVPvHPSetBMCQ \\
    KPrim     & \nVPvHPSetBKPrim & \meanVPvHPSetBKPrimVP & \meanVPvHPSetBKPrimHP & \deltaVPvHPSetBKPrim~$[\ciloVPvHPSetBKPrim, \cihiVPvHPSetBKPrim]$ & \pvalVPvHPSetBKPrim \\
    Free text & \nVPvHPSetBFT    & \meanVPvHPSetBFTVP    & \meanVPvHPSetBFTHP    & \deltaVPvHPSetBFT~$[\ciloVPvHPSetBFT, \cihiVPvHPSetBFT]$         & \pvalVPvHPSetBFT \\
    \bottomrule
  \end{tabular}
\end{table}

%% file: results/no_image_ablation.tex
\begin{table}[h]
  \centering
  \small
  \caption{\textbf{Image ablation across lead models on VIPER.} ``Normal'' is the question's correct image; ``Black'' a solid black image; ``Random'' a randomly-chosen \emph{other} benchmark image; ``No image'' drops the image entirely. Right-most column is the Normal $-$ No-image gap. The Random condition is reported on MCQ only; the KPrim and free-text random-image conditions used an earlier protocol and are omitted for comparability.}
  \label{tab:ablation-matrix}
  \begin{tabular}{lcrrrrr}
    \toprule
    Model & Question type & Normal & Black & Random & No image & $\Delta$ (Normal $-$ No) \\
    \midrule
    \multirow{3}{*}{ToxScribe (Qwen3.5)}
      & MCQ      & 67.1 & 45.1 & 45.8 & 43.2 & $-$24.0 \\
      & KPrim    & 61.8 & 19.2 & --   & 13.2 & $-$48.7 \\
      & Free-text& 58.3 & 11.4 & --   &  5.7 & $-$52.6 \\
    \midrule
    \multirow{3}{*}{PathChat+}
      & MCQ      & 58.7 & 36.3 & 34.1 & 27.5 & $-$31.2 \\
      & KPrim    & 41.5 & 18.4 & --   & 15.0 & $-$26.6 \\
      & Free-text& 52.7 & 10.2 & --   & 26.5 & $-$26.1 \\
    \midrule
    \multirow{3}{*}{GPT-5.4}
      & MCQ      & 58.5 & 34.2 & 33.1 & 30.5 & $-$27.9 \\
      & KPrim    & 54.3 & 31.0 & --   & 25.1 & $-$29.2 \\
      & Free-text& 55.1 & 30.8 & --   & 25.6 & $-$29.5 \\
    \bottomrule
  \end{tabular}
\end{table}

%% file: results/pathmmu.tex
\begin{table}[h]
  \caption{\textbf{Cross-benchmark generalization on PathMMU.} ($n=8{,}468$; 95\% bootstrap CIs, 10k resamples). Both ToxScribe variants show some transfer to human pathology: ToxScribe (Gemma) is statistically similar to GPT-5.4 ($\Delta=-0.7$\,pp, $P=0.20$, paired bootstrap). PathChat+ is the best performer, leading over ToxScribe (Qwen) by 2.1\% ($P<0.001$).}
  \label{tab:pathmmu}
  \centering
  \footnotesize
  \setlength{\tabcolsep}{3pt}
  \begin{tabular}{lcccccc}
    \toprule
    Model & Atlas & EduContent & PubMed & PathCLS & SocialPath & Overall \\
          & $n{=}799$ & $n{=}1{,}683$ & $n{=}2{,}787$ & $n{=}1{,}632$ & $n{=}1{,}567$ & $n{=}8{,}468$ \\
    \midrule
    ToxScribe (Qwen)  & \underline{78.7}\,\tiny{[75.8,81.6]} & 69.1\,\tiny{[66.8,71.2]} & 69.0\,\tiny{[67.2,70.7]} & 42.1\,\tiny{[39.7,44.5]} & 70.3\,\tiny{[68.0,72.5]} & 65.0\,\tiny{[64.0,66.0]} \\
    ToxScribe (Gemma) & 77.6\,\tiny{[74.6,80.5]} & 69.8\,\tiny{[67.6,71.9]} & 72.3\,\tiny{[70.7,74.0]} & \underline{52.1}\,\tiny{[49.7,54.5]} & \underline{70.5}\,\tiny{[68.2,72.7]} & 68.1\,\tiny{[67.1,69.1]} \\
    PathChat+         & \textbf{84.0}\,\tiny{[81.5,86.5]} & \textbf{72.9}\,\tiny{[70.8,75.0]} & \textbf{75.2}\,\tiny{[73.6,76.9]} & 51.3\,\tiny{[48.9,53.7]} & \textbf{71.0}\,\tiny{[68.7,73.3]} & \textbf{70.2}\,\tiny{[69.2,71.2]} \\
    GPT-5.4           & 74.2\,\tiny{[71.2,77.2]} & \underline{70.2}\,\tiny{[67.9,72.3]} & \underline{73.2}\,\tiny{[71.6,74.9]} & \textbf{56.2}\,\tiny{[53.8,58.6]} & 69.8\,\tiny{[67.6,72.1]} & \underline{68.8}\,\tiny{[67.8,69.8]} \\
    \bottomrule
  \end{tabular}
\end{table}

%% file: prompts.tex
\subsection{Prompts}
\label{app:prompts}

\guillaume{Prompts used to augment seed questions into VIPER questions. The generation and adversarial checking both use \texttt{gpt-5.2}. The checker uses \texttt{temperature=0} with no image. \texttt{\{...\}} marks a template field filled per image.}

\textbf{1. Generation, system prompt.}\\

\begin{lstlisting}[style=viperprompt]
You are an expert veterinary pathologist and exam question author. You create high-quality exam questions about histopathology images for a veterinary pathology course.

CONTEXT:
- All images are hematoxylin & eosin (H&E) stained histopathology sections from RATS
- Images come from either TG-GATEs or MMO 
- The target audience is veterinary pathology students and residents

You will be given:
1. A histopathology image
2. A seed question and answer about that image (created by an expert pathologist)
3. Metadata (organ, source, magnification)

Your task: Generate THREE exam question variations based on the seed Q&A and image:

===============================
A. MCQ TYPE A (Single-choice: 1 correct answer + 4 distractors)
===============================
Scoring: 1 point for correct, 0 for a distractor.

QUESTION STEM RULES (violations are NOT accepted):
- Must NOT contain superfluous information or instructions
- Must be clearly formulated and easy to understand
- Must include necessary context (animal species = rat, reference image)
- Must NOT contain unfamiliar abbreviations (spell out on first use)
- Must NOT contain negations (no "which is NOT..." unless absolutely justified)
- Must be answerable directly without seeing the options (Cover-the-Options rule)
- Must be a complete question ending with a question mark
- Must reference the image when relevant

ANSWER OPTION RULES (violations are NOT accepted):
- The correct answer must NOT be identifiable by length alone (all options similar length)
- Options must be homogeneous (same style, structure, and dimension)
- Options must NOT contain word repetition from the stem that hints at the answer
- Options must NOT contain negations
- Options must be concise; move shared information to the stem
- Options must NOT contain absolute/vague statements ("always", "never", "frequently")
- Each option must contain exactly ONE statement
- All options must lie on the SAME dimension (all diagnoses, all structures, etc.)

TECHNICAL CRITERIA:
- Tests relevant veterinary pathology knowledge
- Factually correct and clearly formulated
- Correct answer is the "single best answer"
- Distractors are plausible but clearly wrong to an expert
- Process-appropriate: observation -> diagnosis -> interpretation

===============================
B. KPRIM (4 statements to assess as true or false)
===============================
Scoring: 1 point for 4/4 correct, 0.5 for 3/4, 0 for <=2/4.

QUESTION STEM RULES (violations are NOT accepted):
- Same as MCQ Type A stem rules
- Must be grammatically neutral (no "Which characteristic...")
- Must NOT contain negations

STATEMENT RULES (violations are NOT accepted):
- Each statement must contain exactly ONE assertion
- Statements must NOT contain negations
- Statements must NOT be countable or interdependent
- No statement should be answerable without specialist knowledge (no hidden cues)
- Must NOT contain absolute/vague statements ("always", "all", "never")
- Must be concise; shared information goes in the stem
- All statements must lie on the SAME dimension
- Statements must be homogeneous in style and length

TECHNICAL CRITERIA:
- Statements are correctly matched (clearly true or clearly false)
- Plausible and unambiguous
- Mix of true and false (ideally 2 true + 2 false, at minimum 1 true + 1 false)

===============================
C. FREE TEXT (Open-ended with scoring rubric)
===============================
Scoring: 1 or 2 points with specified distribution.

QUESTION STEM RULES (violations are NOT accepted):
- Same as MCQ Type A stem rules
- Must NOT contain vague expressions ("usual", "frequent", "common")
- Must cover only ONE subtopic (no double questions)
- Must be formulated neutrally (no personal opinions)
- Must precisely specify what and how to answer
- Must have a meaningful scoring rubric

ANSWER RULES (violations are NOT accepted):
- Fair proportion between expected solution, processing time, and points
- Partial points clearly specified
- Solution horizon and evaluation scheme well-defined
- Keywords and sample solution precise and complete
- Provide synonyms that should be accepted as correct

TECHNICAL CRITERIA:
- Expected solution is clear and based on expert consensus
- Synonyms are provided for acceptable alternative phrasings

===============================
ANTI-LEAKAGE -- CRITICAL (applies to ALL question types):
===============================
This is a VISUAL question-answering benchmark. A reader who sees ONLY the question text
(without the image) must perform NO BETTER than random guessing.

1. The question stem must direct students to LOOK AT the image to identify, describe,
   or diagnose -- NOT to recall textbook facts about a finding already named in the stem.
2. Do NOT name or describe findings in the stem that give away the answer.
3. All MCQ options / Kprim statements must be equally plausible a priori for the given organ.
4. All MCQ options must be similar length (+/-20% word count). One option being longer or
   more specific than others is a text-only giveaway.
5. SELF-CHECK: Before finalizing, imagine you can ONLY read the question text with NO
   image. Could you guess the answer? If yes, rewrite until you cannot.

===============================
IMPORTANT: Before finalizing each question, mentally verify it against ALL the rules above.
Output valid JSON matching the required schema exactly.
\end{lstlisting}

\textbf{2. Generation, user prompt.} \\

\begin{lstlisting}[style=viperprompt]
METADATA:
- Source: {source}
- Organ: {organ} (rat)
- Magnification: {magnification}

SEED QUESTION (from expert pathologist):
Q: {seed_question}
A: {seed_answer}

Based on the histopathology image and the seed Q&A above, generate three exam question variations:
1. MCQ Type A (single best answer + 4 distractors)
2. Kprim (4 statements, each true or false)
3. Free text (open question with expected answer, synonyms, and scoring rubric)

CRITICAL: This is a VISUAL question-answering benchmark. Every question MUST require examining the image to arrive at the correct answer. A reader who sees ONLY the question text without the image should perform no better than random guessing.

Key rules:
- The question stem must NOT name or describe findings visible in the image -- the student must identify them.
- Do NOT copy the seed answer verbatim -- use the seed as factual basis but reformulate.
- All MCQ options must be plausible for rat {organ}. All options must be the same length (+/-20% word count).
- For MCQ: after writing all 5 options, verify that WITHOUT the image a pathology expert would consider each option equally likely (~20%). If one option is obviously correct from text alone (e.g., it's the "normal" option among severe pathology distractors), REVISE.
- For Kprim: each statement's truth value must require examining this specific image, not just knowing pathology facts.
\end{lstlisting}

\textbf{3. Structured output schema.}\\

\begin{lstlisting}[style=viperprompt]
class MCQDistractorOut(BaseModel):
    text: str          # The distractor text
    explanation: str   # Why this distractor is wrong but plausible

class MCQTypeAOut(BaseModel):
    stem: str                  # ends with a question mark
    correct_answer: str        # the single best correct answer
    correct_explanation: str
    distractors: list[MCQDistractorOut]   # min_length=4, max_length=4

class KprimStatementOut(BaseModel):
    text: str
    is_true: bool
    explanation: str

class KprimQuestionOut(BaseModel):
    stem: str
    statements: list[KprimStatementOut]   # min_length=4, max_length=4

class FreeTextQuestionOut(BaseModel):
    stem: str
    expected_answer: str
    synonyms: list[str]
    scoring_rubric: str
    max_points: int            # 1 or 2
\end{lstlisting}

{\textbf{4. Adversarial text-only checker, MCQ, system prompt.}} \\

\begin{lstlisting}[style=viperprompt]
You are a veterinary pathology student taking an exam. You are answering a multiple-choice question about a histopathology image, but you CANNOT see the image. You must answer based ONLY on the question text and your general knowledge of veterinary pathology.

Pick the single best answer. If you cannot determine the answer without the image, make your best guess based on the available textual cues.

IMPORTANT: You must choose exactly one option (A, B, C, D, or E). Do not refuse to answer.

Output ONLY a JSON object: {"answer_index": <0-4>, "confidence": "high"|"medium"|"low", "reasoning": "<brief explanation>"}
\end{lstlisting}

\textbf{5. Adversarial text-only checker, MCQ, user prompt.} \\ 

\begin{lstlisting}[style=viperprompt]
Question: {stem}

Options:
  A. {option_1}
  B. {option_2}
  C. {option_3}
  D. {option_4}
  E. {option_5}

Which is the correct answer? Respond with JSON only.
\end{lstlisting}

{\textbf{6. Adversarial text-only checker, KPrim, system prompt.}} \\

\begin{lstlisting}[style=viperprompt]
You are a veterinary pathology student taking an exam. You are evaluating 4 true/false statements about a histopathology image, but you CANNOT see the image. You must answer based ONLY on the statement text and your general knowledge of veterinary pathology.

For each statement, judge whether it is true or false. If you cannot determine the answer without the image, make your best guess based on textual cues.

IMPORTANT: You must provide a true/false judgment for all 4 statements. Do not refuse to answer.

Output ONLY a JSON object: {"answers": [true|false, true|false, true|false, true|false], "confidence": "high"|"medium"|"low", "reasoning": "<brief explanation>"}
\end{lstlisting}

\textbf{7. Adversarial text-only checker, KPrim, user prompt.}

\begin{lstlisting}[style=viperprompt]
Question: {stem}

Statements:
  1. {statement_1}
  2. {statement_2}
  3. {statement_3}
  4. {statement_4}

For each statement, decide if it is true or false based on your pathology knowledge. Respond with JSON only.
\end{lstlisting}

\textbf{8. Regeneration after an adversarial failure, MCQ, user prompt (system prompt unchanged from item 1).} \\

\begin{lstlisting}[style=viperprompt]
METADATA:
- Source: {source}
- Organ: {organ} (rat)
- Magnification: {magnification}

SEED QUESTION: Q: {seed_question} / A: {seed_answer}

YOUR PREVIOUS MCQ WAS REJECTED because a text-only reader (without the image) guessed the correct answer.

FAILED QUESTION:
Stem: {failed_stem}
Correct: {failed_correct_answer}
Distractors:
  - {failed_distractor_1}
  - {failed_distractor_2}
  - {failed_distractor_3}
  - {failed_distractor_4}

WHY IT WAS GUESSABLE: {checker_reasoning}

Generate a NEW MCQ that fixes this problem. The text-only reader must NOT be able to pick the correct answer.

CRITICAL CONSTRAINT -- SEED FIDELITY:
The new question MUST still test the same knowledge as the seed question/answer. The correct answer must remain grounded in the seed answer. Do NOT change the topic. Instead, change HOW you ask about the same topic.

Strategies to stay on-topic while avoiding guessability:
- Ask about the VISUAL APPEARANCE or SPATIAL LOCATION of the finding described in the seed (e.g., "Where in the image is X located?" or "What does X look like in this section?").
- Ask about the DISTRIBUTION PATTERN, COUNT, or EXTENT of the seed finding.
- Ask about the RELATIONSHIP between the seed finding and adjacent structures visible in the image.
- Frame the question so the correct answer describes an image-specific visual detail of the seed finding, not just its name.
- Make ALL distractors describe equally plausible visual appearances, locations, or patterns for the same general class of finding.
- If the reader guessed because one option was "the textbook answer", make ALL options equally common/typical while keeping them relevant to the seed topic.
- If the reader guessed from the stem hinting at the finding, remove diagnostic terms from the stem but keep it on-topic.
- If one option had more detail or sounded more specific, equalize all options.

All 5 options must be the same length (+/-20% word count). Reference the image.
\end{lstlisting}

\textbf{9. Regeneration after an adversarial failure, KPrim, user prompt (system prompt unchanged from item 1).} \\

\begin{lstlisting}[style=viperprompt]
METADATA:
- Source: {source}
- Organ: {organ} (rat)
- Magnification: {magnification}

SEED QUESTION: Q: {seed_question} / A: {seed_answer}

YOUR PREVIOUS KPRIM QUESTION WAS REJECTED because a text-only reader (without the image) correctly guessed {n_correct}/4 statements.

These statements were guessable:
  Statement {i} ({true|false}): "{statement_text}"

WHY THEY WERE GUESSABLE: {checker_reasoning}

CRITICAL CONSTRAINT -- SEED FIDELITY:
The new Kprim question MUST still test the same topic as the seed question/answer. All four statements must remain grounded in the seed topic. Do NOT change the subject.

Generate a NEW Kprim question where EACH statement's truth value can ONLY be determined by looking at the image.
Strategies to stay on-topic while avoiding guessability:
- Make statements about the SPATIAL LOCATION, DISTRIBUTION, or VISUAL APPEARANCE of the finding from the seed.
- Use statements about COUNTS, PROPORTIONS, or RELATIONSHIPS between the seed finding and adjacent structures.
- Replace guessable "is X present?" statements with "X is located in [specific region]" or "X shows [specific visual pattern]" -- these require image inspection.
- Avoid statements that are generally true/false for this organ -- make them specific to THIS image.
- Each statement should describe something that COULD be true or false depending on what this specific image shows.
\end{lstlisting}

\textbf{10. Regeneration after a pathologist rejection, system prompt.} \\

\begin{lstlisting}[style=viperprompt]
You are an expert veterinary pathologist and exam question author. You create high-quality exam questions about histopathology images for a veterinary pathology course.

CONTEXT:
- All images are hematoxylin & eosin (H&E) stained histopathology sections from RATS
- Images come from either TG-GATEs (liver toxicogenomics) or MMO (multi-organ study: liver, kidney)
- The target audience is veterinary pathology students and residents

You will be given:
1. A histopathology image
2. A seed question and answer about that image (created by an expert pathologist)
3. Metadata (organ, source, magnification)
4. A previously generated question that was rejected by a reviewer, along with their feedback
\end{lstlisting}

\textbf{11. Regeneration after a pathologist rejection, user prompt.} \\

\begin{lstlisting}[style=viperprompt]
METADATA:
- Source: {source}
- Organ: {organ} (rat)
- Magnification: {magnification}

SEED QUESTION (from expert pathologist):
Q: {seed_question}
A: {seed_answer}

PREVIOUS {MCQ TYPE A | KPRIM | FREE TEXT} QUESTION (REJECTED BY REVIEWER):
{rejected_question_rendered}

REVIEWER FEEDBACK: {reviewer_note}

OTHER EXISTING QUESTIONS FOR THIS IMAGE (for context -- ensure your new question tests a DIFFERENT aspect):
{other_two_formats_rendered}

Generate an improved {MCQ Type A | Kprim | Free Text} question that addresses the reviewer's feedback.
Do NOT repeat the same approach as the rejected question -- generate a meaningfully different and better variation.
All questions must reference this histopathology image of rat {organ}.
\end{lstlisting}